\documentclass[10pt,letterpaper]{article}
\usepackage[top=0.85in,left=2.75in,footskip=0.75in]{geometry}

\usepackage[normalem]{ulem}

\usepackage{amsmath,amssymb}

\usepackage{changepage}

\usepackage[utf8x]{inputenc}

\usepackage{textcomp,marvosym}

\usepackage{cite}

\usepackage{nameref,hyperref}

\usepackage[right]{lineno}

\usepackage{microtype}
\DisableLigatures[f]{encoding = *, family = * }

\usepackage[table]{xcolor}

\usepackage{array}

\newcolumntype{+}{!{\vrule width 2pt}}

\newlength\savedwidth

\raggedright
\usepackage[aboveskip=1pt,labelfont=bf,labelsep=period,justification=raggedright,singlelinecheck=off]{caption}

\makeatletter
\renewcommand{\@biblabel}[1]{\quad#1.}
\makeatother

\usepackage{lastpage,fancyhdr,graphicx}
\usepackage{epstopdf}
\fancyheadoffset[L]{2.25in}
\fancyfootoffset[L]{2.25in}
\newcommand{\side}[1]{\begin{sideways}{#1}\end{sideways}}

\newif\ifdraft
\drafttrue

\usepackage{rotating}
\newcommand{\pdag}{\phantom{$^{\dag}$}}

\newcommand{\fabsebcomment}[1]{\ifdraft{\leavevmode\color{blue}{[FS]: {#1}}}\else{\vspace{0ex}}\fi}

\newcommand{\blue}[1]{\ifdraft{\leavevmode\color{black}{#1}}\else{\leavevmode\color{black}{#1}}\fi}

\newcommand{\MLPE}{\mathrm{MLPE}}

\newcommand{\EPACCPtr}{E(PACC)$_{\mathrm{Ptr}}$}
\newcommand{\EPACCAE}{E(PACC)$_{\mathrm{AE}}$}

\DeclareMathOperator{\aae}{AE}

\DeclareMathOperator{\rae}{RAE}

\newcolumntype{.}{D{.}{.}{-1}}
\newcolumntype{M}[1]{>{\centering\arraybackslash}m{#1}}
\newcolumntype{N}{@{}m{0pt}@{}}
\newcolumntype{C}[1]{>{\centering\let\newline\\\arraybackslash\hspace{0pt}}m{#1}}
\newcolumntype{Y}{>{\centering\arraybackslash}X}

\begin{document}
\vspace*{0.2in}

\begin{flushleft}
{\Large
\textbf\newline{Tweet Sentiment Quantification: \\ An Experimental
Re-Evaluation} 
}
\newline
\\
Alejandro Moreo\textsuperscript{1\ddag*},
Fabrizio Sebastiani\textsuperscript{1\ddag}
\\
\bigskip
\textbf{1} Istituto di Scienza e Tecnologie dell'Informazione, Consiglio Nazionale delle Ricerche, 56124 Pisa, Italy
\\
\bigskip

%
%
\ddag The order in which the authors are listed is purely alphabetical; each author has given an equally important contribution to this work.





* alejandro.moreo@isti.cnr.it

\end{flushleft}

\begin{abstract}
  \noindent Sentiment quantification is the task of \blue{training, by
  means of supervised learning, estimators of} the relative frequency
  (\blue{also called} ``prevalence'') of sentiment-related classes
  (such as \textsf{Positive}, \textsf{Neutral}, \textsf{Negative}) in
  a sample of unlabelled texts. \blue{This task} is especially
  important when these texts are tweets, since \blue{the final goal of
  most sentiment classification efforts carried out on Twitter data is
  actually quantification (and not the classification of individual
  tweets)}. It is well-known that solving quantification \blue{by
  means of} ``classify and count'' (i.e., by classifying all
  unlabelled items \blue{by means of} a standard classifier and
  counting the items that have been assigned to a given class) is
  \blue{less than optimal} in terms of accuracy, and that more
  accurate quantification methods exist.  \blue{Gao and Sebastiani~\cite{Gao:2016uq}}
  carried out a systematic comparison of quantification methods on the
  task of tweet sentiment quantification. In hindsight, we observe
  that the experimental protocol followed in that work was
  \blue{weak}, and that \blue{the reliability of the conclusions that
  were drawn from the results is thus questionable}. We now
  re-evaluate those quantification methods (plus a few more modern
  ones) on \blue{exactly the same} same datasets, this time following
  a now consolidated and much more robust experimental protocol
  \blue{(which also involves simulating the presence, in the test
  data, of class prevalence values very different from those of the
  training set)}.  \blue{This experimental protocol} (even without
  counting the newly added methods) involves \blue{a number of
  experiments 5,775 times larger than that of the original
  study}. \blue{The results of our experiments are} dramatically
  different from those obtained by Gao and Sebastiani, and \blue{they}
  provide a different, much more solid understanding of the relative
  strengths and weaknesses of different sentiment quantification
  methods.
\end{abstract}




\section{Introduction}
\label{sec:intro}

\noindent 
\emph{Quantification} (\blue{also known as} \emph{supervised
prevalence estimation}, or \emph{learning to quantify}) is the task of
training \blue{(by means of supervised learning)} a predictor that
estimates the relative frequency (\blue{also known as}
\emph{prevalence}, or \emph{prior probability}) of the classes of
interest in a \blue{set (here often called a ``sample'')} of
unlabelled data items, where the data used to train the predictor are
a set of labelled data items
\cite{Gonzalez:2017it}.\footnote{\label{foot:unlabelled}\blue{Throughout
the paper we prefer the term ``unlabelled text'' to the term ``test
text'' because the former embraces not only the case in which we are
testing a quantification method in lab experiments, but also the case
in which (maybe after performing these experiments) we deploy our
quantifiers in an operational environment in order to perform
quantification on the data that our application requires us to
analyse.} } Quantification finds applications in fields (such as the
social sciences \cite{Hopkins:2010fk}, epidemiology
\cite{King:2008fk}, market research \cite{Esuli:2010kx}, and
ecological modelling \cite{Beijbom:2015yg}) that inherently deal with
aggregate (rather than individual) data, but is also relevant to other
applications such as resource allocation \cite{Forman:2008kx}, word
sense disambiguation \cite{Chan2006}, and improving classifier
fairness \cite{Biswas:2019vn}.

In the realm of textual data, one important domain to which
quantification is applied is \emph{sentiment
analysis}~\blue{\cite{Liu:2012zk,Pang08}}. In fact, as argued
by Esuli et al.~\cite{Esuli:2010fk}, many applications of sentiment classification
are such that the final goal is not determining the class label (e.g.,
\textsf{Positive}, or \textsf{Neutral}, or \textsf{Negative}) of an
individual \blue{unlabelled} text (\blue{for example}, a blog post, a
response to an open question, or a comment on a product), but is that
of determining the relative frequencies of the classes of interest in
a set of unlabelled texts. In a 2016 paper, Gao and Sebastiani~\cite{Gao:2016uq}
(hereafter: [GS2016]) have argued that, when the objects of analysis
are tweets, the \emph{vast majority} of sentiment classification
\blue{efforts} actually have quantification as their final goal, since
hardly anyone who engages in sentiment classification of tweets is
interested in the sentiment conveyed by a specific tweet. \blue{We
call the resulting task \emph{tweet sentiment
quantification}~\blue{\cite{Ayyub:2020tm,Esuli:2010fk,Gao:2016uq}}.
}

It is well-known \blue{(see e.g.,
\cite{Fiksel:2021tw,Forman:2005fk,Forman:2008kx,Gonzalez:2017it,Hassan:2020kq,Maletzke:2019qd,Perez-Gallego:2017wt,Qi:2020qp,Schumacher:2021ty,Tasche:2021df})}
that solving quantification \blue{by means of} ``classify and count''
(i.e., by classifying all \blue{the} unlabelled items \blue{by means
of} a standard classifier and counting the items that have been
assigned to a given class) is \blue{less than optimal} in terms of
accuracy, and that more accurate quantification methods exist.
Driven by these considerations, [GS2016] presented an experimental
comparison of 8 important quantification methods on 11 Twitter
datasets annotated by sentiment, with the goal of assessing the
strengths and weaknesses of the various methods for tweet sentiment
quantification. That paper became then influential\footnote{At the
time of writing, [GS2016] and paper \cite{Gao:2015ly} (a shorter and
earlier version of [GS2016]) have \blue{118} citations altogether on
Google Scholar.} and a standard reference on this problem, and
describes what is \blue{currently} the largest comparative
experimentation on \blue{tweet} sentiment quantification.

In this paper we argue that the experimental results obtained in
[GS2016] are unreliable, as a result of the fact that the experimental
protocol used in that paper was \blue{weak}. We thus present new
experiments in which we re-test all 8 quantification methods
originally tested in [GS2016] (plus some additional ones that have
been proposed since then) on the same 11 datasets used in [GS2016],
this time using a now consolidated and much more robust experimental
protocol. These new experiments (\blue{whose number is} \blue{5,775}
times larger than \blue{the number of experiments conducted in}
[GS2016], even without counting the experiments on new quantification
methods that had not been considered in [GS2016]) \blue{return}
results dramatically different from those obtained in [GS2016], and
thus give us a new, more reliable picture of the relative merits of
the various methods on the tweet sentiment quantification task.

The rest of this paper is structured as follows. In
Section~\ref{sec:protocols} we discuss experimental protocols for
quantification, and argue why \blue{the experimentation carried out}
in [GS2016] is, in hindsight, \blue{weak}.  In
Section~\ref{sec:experiments} we present the new experiments we have
run, briefly discussing the quantification methods and the datasets we
use, and explaining in detail the experimental protocol we
use. Section~\ref{sec:results} discusses the results and the
conclusions that they allow \blue{drawing}, also pointing at how they
differ from the ones of [GS2016].  Section~\ref{sec:conclusions} is
devoted to concluding remarks.

We make all the code we use for our experiments
available.\footnote{\url{https://github.com/HLT-ISTI/QuaPy/tree/tweetsent}}
Together with the fact that~[GS2016] made available (in vector form)
all their 11
datasets\footnote{\label{foot:datasets}\url{https://zenodo.org/record/4255764}},
this allows our experiments to be easily reproduced by other
researchers.


\section{Experimental Protocols for Quantification}
\label{sec:protocols}


\subsection{Notation}
\label{sec:notation}

\noindent In this paper we use the following notation.  By
$\mathbf{x}$ we indicate a document drawn from a domain $\mathcal{X}$
of documents, while by $y$ we indicate a class drawn from a set of
classes (\blue{also known as a} \emph{codeframe})
$\mathcal{Y}=\{y_{1}, ..., y_{|\mathcal{Y}|}\}$.
Given $\mathbf{x}\in\mathcal{X}$ and $y\in\mathcal{Y}$, a pair
$(\mathbf{x},y)$ thus denotes a document with its true class label.
Symbol $\sigma$ denotes a \index{Sample}\emph{sample}, i.e., a
non-empty set of (labelled or unlabelled) documents drawn from
$\mathcal{X}$.
By $p_{\sigma}(y)$ we indicate the true prevalence of class $y$ in
sample $\sigma$, by $\hat{p}_{\sigma}(y)$ we indicate an estimate of
this prevalence\footnote{Consistently with most mathematical
literature, we use the caret symbol (\^\/\/) to indicate estimation.},
and by $\hat{p}_{\sigma}^{M}(y)$ we indicate the estimate of this
prevalence obtained \blue{by means of}
quantification method $M$.
\blue{Since $0\leq p_{\sigma}(y) \leq 1$ and
$0\leq \hat{p}_{\sigma}(y) \leq 1$ for all $y\in\mathcal{Y}$, and
since
$\sum_{y\in\mathcal{Y}}p_{\sigma}(y)=\sum_{y\in\mathcal{Y}}\hat{p}_{\sigma}(y)=1$,
the $p_{\sigma}(y)$'s and the $\hat{p}_{\sigma}(y)$'s form two
probability distributions across the same codeframe.}

By $D(p,\hat{p})$ we denote an evaluation measure for quantification;
these measures are typically \emph{divergences}, i.e., functions that
measure the amount of discrepancy between two \blue{probability}
distributions. By $L$ we denote a \blue{set} of \underline{l}abelled
documents, that we typically use as a training set, while by $U$ we
denote a \blue{set} of \underline{u}nlabelled documents, that we
typically use as a sample to quantify on.  We take a \emph{hard
classifier} to be a function $h:\mathcal{X}\rightarrow \mathcal{Y}$,
and a \emph{soft classifier} to be a function
$s:\mathcal{X}\rightarrow [0,1]^{|\mathcal{Y}|}$, where
$s(\mathbf{x})$ is a vector of $|\mathcal{Y}|$ \emph{posterior
probabilities} \index{Probability!posterior}(each indicated as
$\Pr(y|\mathbf{x})$), such that
$\sum_{y\in \mathcal{Y}}\Pr(y|\mathbf{x})=1$; $\Pr(y|\mathbf{x})$
indicates the probability of membership in $y$ of item $\textbf{x}$ as
estimated by the soft classifier $s$. By $\delta_{\sigma}(y)$ we
denote the set of documents in sample $\sigma$ that have been assigned
to class $y$ by a hard classifier.


\subsection{Why do we need quantification?}
\label{sec:why}

\noindent Quantification may be seen as the task of approximating a
\emph{true distribution} $p_{\sigma}$, where $p_{\sigma}$ is defined
\blue{on a sample $\sigma$ and} across the classes in a codeframe
$\mathcal{Y}=\{y_{1}, ..., y_{|\mathcal{Y}|}\}$, by means of a
\emph{predicted distribution} $\hat{p}_{\sigma}$; in other words, in
quantification one needs to generate estimates
$\hat{p}_{\sigma}(y_{1}), ..., \hat{p}_{\sigma}(y_{|\mathcal{Y}|})$ of
the true (and unknown) class prevalence values
$p_{\sigma}(y_{1}), ..., p_{\sigma}(y_{|\mathcal{Y}|})$, where
$\sum_{y\in\mathcal{Y}} \hat{p}_{\sigma}(y) =
\sum_{y\in\mathcal{Y}}p_{\sigma}(y)=1$. In this paper we consider a
ternary sentiment quantification task (an example of
\emph{single-label multiclass quantification}) in which the codeframe
is
$\mathcal{Y}=\{\textsf{Positive}, \textsf{Neutral},
\textsf{Negative}\}$, \blue{and} where these three class labels will
be indicated, for brevity, by the symbols
$\{\oplus, \odot, \ominus\}$. All the 11 datasets discussed in
Section~\ref{sec:datasets} use this codeframe.

The reason why true quantification methods (i.e., different from the
trivial ``classify and count'' mentioned in Section~\ref{sec:intro})
are needed is the fact that many applicative scenarios suffer from
\emph{distribution shift}, the phenomenon according to which the
distribution $p_{L}(y)$ in the training set $L$ may substantially
differ from the distribution $p_{U}(y)$ in the unlabelled data $U$
that one needs to label
\blue{\cite{Moreno-Torres:2012ay,Quinonero09}}.
The presence of distribution shift means that the well-known IID
assumption, on which most learning algorithms for training classifiers
\blue{are based}, does not hold; in turn, this means that ``classify
and count'' will perform \blue{less than optimally} on sets of
unlabelled items that exhibit distribution shift with respect to this
training set, and that the higher the amount of shift, the worse we
can expect ``classify and count'' to perform.


\subsection{The APP and the NPP}
\label{sec:appnpp}

\noindent
There are two main experimental protocols that have been used in the
literature for evaluating quantification; we will here call them the
\emph{artificial-prevalence protocol} (APP) and the
\emph{natural-prevalence protocol} (NPP).

The APP consists of taking a standard dataset\footnote{\blue{By ``a
standard dataset'' we here mean any dataset that has originally been
assembled for testing classification systems; any such dataset can be
used for testing quantification systems too.}}, split into a training
set $\mathcal{L}$ of labelled items and a set $\mathcal{U}$ of
unlabelled items, and conducting repeated experiments in which either
the training set prevalence values or the test set prevalence values
of the classes are artificially varied \blue{by means of} subsampling
(i.e., by removing random elements of \blue{specific} classes until
the desired class prevalence values are obtained). \blue{In other
words, subsampling is used either to generate $s$ training samples
$L_{1}\subseteq\mathcal{L}, ..., L_{s}\subseteq\mathcal{L}$, or to
generate $t$ test samples
$U_{1}\subseteq\mathcal{U}, ..., U_{t}\subseteq\mathcal{U}$, or both,
where the class prevalence values of the generated samples are
predetermined and set in such a way as to generate a wide array of
distribution drift values.} This is meant to test the robustness of a
\emph{quantifier} (i.e., of an estimator of class prevalence values)
in \blue{scenarios characterized by class prevalence} values very
different from the ones \blue{the quantifier} has been trained on.
For instance, in the binary quantification experiments carried out in
\cite{Forman:2005fk}, given codeframe $\mathcal{Y}=\{y_{1},y_{2}\}$,
repeated experiments are conducted in which examples of either $y_{1}$
or $y_{2}$ are removed at random from the test set in order to
generate predetermined prevalence values for $y_{1}$ and $y_{2}$ in
the \blue{samples $U_{1}, ..., U_{t}$} thus obtained. In this way,
\blue{the different samples are characterised} by a different
prevalence of $y_{1}$ (e.g.,
$p_{U}(y_{1})\in \{0.00, 0.05, ..., 0.95, 1.00\}$) and, as a result,
by a different prevalence of $y_{2}$. This can be repeated, thus
generating multiple random samples for each chosen pair of class
prevalence values. Analogously, random removal of examples of either
$y_{1}$ or $y_{2}$ can be performed on the training set, thus bringing
about training \blue{samples} with different values of $p_{L}(y_{1})$
and $p_{L}(y_{2})$.

This protocol had been criticised (see \cite{Esuli:2015gh}) because
it may generate samples exhibiting class prevalence values very
different from the ones of the set from which the sample was
extracted, i.e., class prevalence values that might be hardly
plausible \blue{in practice}. As a result, one may resort to the NPP,
which consists instead of conducting experiments on ``real'' datasets
only, i.e., datasets consisting of a training set $L$ and a test set
$U$ that have been sampled IID from the data distribution. \blue{In
other words, no extraction of samples from the dataset is performed by
perturbing the original class prevalence values; instead, a single
train-and-test run is performed, using the original training set
$\mathcal{L}$ as the training sample $L$ and the original test set
$\mathcal{U}$ as the test sample $U$.}

The experimentation conducted by~[GS2016] on tweet sentiment
quantification is indeed an example of the NPP, since it relies on 11
``original'' datasets of tweets annotated by sentiment, i.e., no
extraction of samples at prespecified values of class prevalence was
performed.  However,~[GS2016] probably failed to
realise that, while in classification an experiment involving 11
different datasets probably counts as large and robust, this does not
hold in quantification \blue{if only one test per dataset is
conducted}. The reason is that, since the objects of quantification
are \emph{sets} of documents \fabsebcomment{in the same way that} the
objects of classification are individual documents, \emph{testing a
quantifier on just 11 sets of documents \blue{should be considered,
from an experimental point of view, a drastically insufficient
experimentation, akin to} testing a classifier on 11 documents only.}

Unfortunately, finding a large enough set \blue{(say, 1,000 or more)}
of datasets sampled IID from the respective data distributions is
\blue{nearly impossible}; this indicates that \blue{extracting a large
enough number of samples from the same dataset} is probably the only
way to go for evaluating quantification.\footnote{An example set of
experiments that use the NPP on a large enough set of test sets is the
one reported in \cite{Esuli:2015gh}, where the authors test
quantifiers on $52\times 99$=\blue{5,148} binary test sets. This
results from the fact that, in using the RCV1-v2 test collection, they
consider the 99 RCV1-v2 classes and bin the RCV1-v2 \blue{791,607}
test documents in 52 bins (each corresponding to a week's worth of
data, since the RCV1-v2 data span one year) of 15,212 documents each
on average. However, it is not always easy to find test collections
with such a large amount of classes and annotated data, and this
limits the applicability of the NPP. It should also be mentioned that,
as Card and Smith~\cite{Card:2018pb} noted, the vast majority of the \blue{5,148}
RCV1-v2 binary test sets used in \cite{Esuli:2015gh} exhibit very
little distribution shift, which makes the testbed used in
\cite{Esuli:2015gh} unchallenging for quantification methods.}
Indeed, most recent quantification works~(e.g.,
\blue{\cite{Esuli:2018rm,Maletzke:2019qd,Perez-Gallego:2017wt,Card:2018pb,Reis:2018fk,Hassan:2020kq,Perez-Gallego:2019vl,Schumacher:2021ty,Vaz:2019eu}})
adopt the APP, and not the NPP.

As a result, we should conclude that the experimentation conducted in
[GS2016] is \blue{weak}, and that the results of that experimentation
are thus unreliable. We thus re-evaluate the same quantification
methods that~[GS2016] tested (plus some other more recent ones) on the
same datasets, this time following the by now consolidated and much
more robust APP; in our case, this turns out to involve \blue{5,775}
as many experiments as run in the original study, even without
considering the experiments on quantification methods that had not
been considered in [GS2016]).

It might be argued that the APP is unrealistic because it generates
test samples whose class prevalence values are too
far away from the values seen in the test set
\blue{from where they have been extracted}, and that such scenarios
are thus unlikely to occur in real applicative
\blue{settings}. However, in the absence of any prior knowledge about
how the class prevalence values are allowed or expected to change in
future data, the APP turns out to be not only the fairest protocol,
since it relies on no assumptions that could penalize or benefit any
particular method, but also the most interesting for quantification,
since quantification is especially useful in cases of distribution
shift.\footnote{Yet another way of saying this comes from the
observation that, should we adopt the NPP instead of the APP, a method
that trivially returns, as the class prevalence estimates for
\blue{\textit{every}} test sample, the class prevalence values from
the training set (this trivial method is commonly known in the
\blue{quantification} literature as the \textit{maximum likelihood
prevalence estimator} -- $\MLPE$), would probably \blue{perform} well,
and might even beat all genuinely engineered quantification
methods. The reason why it would probably \blue{perform} well is that
the expectations of the class prevalence values of samples drawn IID
from the test set coincide with the class prevalence values of the
test set, \blue{and these}, again by virtue of the IID assumption, are
likely to be close to those of the training set. In other words, the
reason why MLPE typically \blue{performs} well when evaluated
according to the NPP, does not lie in the (inexistent) qualities of
MLPE as a quantification method, but in the fact that the NPP is a
weak evaluation protocol.
}


\section{Experiments}
\label{sec:experiments}

\noindent In this section \blue{describe} the experiments we have
carried out in order to re-assess the merits of different
quantification methods under the lens of the APP. \blue{We have
conducted all these} experiments using
QuaPy\footnote{\url{https://github.com/HLT-ISTI/QuaPy}}, a software
framework for quantification written in Python that we have developed
and made available \blue{through} GitHub.\footnote{Please see branch
\texttt{tweetsent}}



\subsection{Evaluation measures}
\label{sec:measures}

\noindent As the measures of quantification error we use
\emph{Absolute Error} ($\aae$) and \emph{Relative Absolute Error}
($\rae$), defined as
\begin{align}
  \label{eq:ae}
  \aae(p,\hat{p}) & =\frac{1}{|\mathcal{Y}|}\sum_{y\in 
                    \mathcal{Y}}|\hat{p}(y)-p(y)| \\
  \label{eq:rae}
  \rae(p,\hat{p}) & =\frac{1}{|\mathcal{Y}|}\sum_{y\in 
                    \mathcal{Y}}\displaystyle\frac{|\hat{p}(y)-p(y)|}{p(y)} 
\end{align}
\noindent where $p$ is the true distribution, $\hat{p}$ is the
estimated distribution, and $\mathcal{Y}$ is the set of classes of
interest ($\mathcal{Y}=\{\oplus,\odot,\ominus\}$ in our case).

Note that $\rae$ is undefined when at least one of the classes
$y\in \mathcal{Y}$ is such that its prevalence in the sample $U$ is
$0$. To solve this problem, in computing $\rae$ we smooth all $p(y)$'s
and $\hat{p}(y)$'s \blue{by means of} additive smoothing, i.e., we
\blue{compute}
%
\begin{align}
  \label{eq:smoothing}
  \underline{p}(y)=\frac{\epsilon+p(y)}{\epsilon|\mathcal{Y}| + 
  \displaystyle\sum_{y\in 
  \mathcal{Y}}p(y)}
\end{align}
\noindent
where $\underline{p}(y)$ denotes the smoothed version of $p(y)$ and
the denominator is just a normalising factor (same for the
$\hat{\underline{p}}(y)$'s); following \cite{Forman:2008kx}, we use
the quantity $\epsilon=1/(2|U|)$ as the smoothing factor. We then use
the smoothed versions of $p(y)$ and $\hat{p}(y)$ in place of their
original non-smoothed versions \blue{in} Equation~\ref{eq:rae}; as a
result, $\rae$ is now always defined.

The reason why we use $\aae$ and $\rae$ is
that from a theoretical standpoint they are, as it has been recently
argued \cite{Sebastiani:2020qf}, the most satisfactory evaluation
measures for quantification. This means that we do not consider other
measures used in [GS2016], such as KLD, NAE, NRAE, and NKLD, since
\cite{Sebastiani:2020qf} shows them to be \blue{inadequate for
evaluating quantification}.


\subsection{Quantification methods used in [GS2016]}
\label{sec:methods}

\noindent We now briefly describe the quantification methods used in
[GS2016], that we also use in this paper.

The simplest quantification method (and the one that acts as a
lower-bound baseline for all quantification methods) is the
above-mentioned \emph{Classify and Count} (\textbf{CC}), which, given
a hard classifier $h$, \blue{consists of} computing
\begin{equation}\label{eq:CC}
  \begin{aligned}
    \hat{p}_{U}^{\mathrm{CC}}(y_{i}) & = \frac{|\{\mathbf{x}\in
    U|h(\mathbf{x})=y_{i}\}|}{|U|} =
    \frac{\sum_{y_{j}\in\mathcal{Y}}C^{h}_{ij}}{|U|}
  \end{aligned}
\end{equation}
\noindent where $C^{h}_{ij}$ indicates the number of documents
classified as $y_{i}$ by $h$ and whose true label is $y_{j}$. CC is an
example of an \emph{aggregative} quantification method, i.e., a method
that requires the (hard or soft) classification of all the unlabelled
items as an intermediate step.  All the methods discussed in this
section
are aggregative.

The \emph{Adjusted Classify and Count} (\textbf{ACC}) quantification
method (see \cite{Forman:2008kx,Vaz:2019eu}) derives from the
observation that, by the law of total probability, it holds that
\begin{align}
  \label{eq:ACC} 
  \Pr(\delta(y_{i})) = \sum_{y_{j}\in
  \mathcal{Y}}\Pr(\delta(y_{i})|y_{j})\cdot \Pr(y_{j})
\end{align}
\noindent where $\delta(y_{i})$ denotes (see
Section~\ref{sec:notation}) the set of documents that have been
assigned to class $y_{i}$ by the hard classifier
$h$. Equation~\ref{eq:ACC} can be more conveniently rewritten as
\begin{align}
  \label{eq:ACC2} 
  \frac{\sum_{y_{j}\in\mathcal{Y}}C^{h}_{ij}}{|U|} = \sum_{y_{j}\in
  \mathcal{Y}}\frac{C^{h}_{ij}}{\sum_{y_{x}\in\mathcal{Y}}C^{h}_{xj}} 
  \cdot p_{U}(y_{j})
\end{align}
\noindent Note that the leftmost factor of Equation~\ref{eq:ACC2} is
known (it is the fraction of documents that the classifier has
assigned to class $y_{i}$, i.e., $\hat{p}_{U}^{\mathrm{CC}}(y_{i})$),
and that $C^{h}_{ij}/\sum_{y_{x}\in\mathcal{Y}}C^{h}_{xj}$ (which
represents the disposition of the classifier to assign $y_{i}$ when
$y_{j}$ \blue{is the true label}), while unknown, can be estimated by
$k$-fold \blue{cross-validation} on $L$. Note also that $p_{U}(y_{j})$
is unknown (it is the goal of quantification to estimate it), and that
there are $|\mathcal{Y}|$ instances of Equation~\ref{eq:ACC}, one for
each $y_{i}\in\mathcal{Y}$. We are then in the presence of a system of
$|\mathcal{Y}|$ linear equations in $|\mathcal{Y}|$ unknowns (the
$p_{U}(y_{j})$'s); ACC thus consists of estimating these latter (i.e.,
computing $\hat{p}_{U}^{\mathrm{ACC}}(y_{j})$) by solving, \blue{by
means of} the known techniques, this system of linear equations.

CC and ACC use the predictions generated by the hard classifier $h$,
as evident by the fact that both Equations~\ref{eq:CC} and
\ref{eq:ACC2} depend on factors of type
$C^{h}_{ij}$. Since most classifiers can be configured to
\blue{return} ``soft predictions'' in the form of posterior
probabilities $\Pr(y|\mathbf{x})$ (from which hard predictions are
obtained by choosing the $y$ for which $\Pr(y|\mathbf{x})$ is
maximised),\footnote{If a classifier natively outputs classification
scores that are not probabilities, the former can be converted into
the latter \blue{by means of} ``probability calibration''; see e.g.,
\cite{Platt:2000fk}.} and since posterior probabilities contain
richer information than hard predictions, it makes sense to try and
generate probabilistic versions of the CC and ACC methods
\cite{Bella:2010kx} by replacing ``hard'' counts $C^{h}_{ij}$ with
their expected values, i.e., with
$C^{s}_{ij}=\sum_{(\mathbf{x},y_{j})\in U}\Pr(y_{i}|\mathbf{x})$.
One can thus define \emph{Probabilistic Classify and Count}
(\textbf{PCC}) as
\begin{equation}\label{eq:PCC}
  \begin{aligned}
    \hat{p}_{U}^{\mathrm{PCC}}(y_{i}) & = \frac{\sum_{\mathbf{x}\in
    U}\Pr(y_{i}|\mathbf{x})}{|U|} =
    \frac{\sum_{y_{j}\in\mathcal{Y}}C^{s}_{ij}}{|U|}
  \end{aligned}
\end{equation}
\noindent and \emph{Probabilistic Adjusted Classify and Count}
(\textbf{PACC}), which consists of estimating $p_{U}(y_{j})$ (i.e.,
computing $\hat{p}_{U}^{\mathrm{PACC}}(y_{j})$) by solving the
\blue{system of $|\mathcal{Y}|$ linear equations in $|\mathcal{Y}|$
unknowns}
\begin{align}
  \label{eq:PACC} 
  \frac{\sum_{y_{j}\in\mathcal{Y}}C^{s}_{ij}}{|U|} = \sum_{y_{j}\in
  \mathcal{Y}}\frac{C^{s}_{ij}}{\sum_{y_{x}\in\mathcal{Y}}C^{s}_{xj}} \cdot p_{U}(y_{j})
\end{align}  
\noindent The fact that PCC is a probabilistic version of CC is
evident from the \blue{structural} similarity between
Equations~\ref{eq:CC} and~\ref{eq:PCC}, which only differ for the fact
that the hard classifier $h$ of Equation~\ref{eq:CC} is replaced by a
soft classifier $s$ in Equation~\ref{eq:PCC}; the same goes for ACC
and PACC, as evident from the structural similarity of
Equations~\ref{eq:ACC2} and~\ref{eq:PACC}.

A further method that [GS2016] uses is the one proposed in
\cite{Saerens:2002uq} (which we here call \textbf{SLD}, from the
names of its proposers, and which was called EMQ in [GS2021]), which
consists of training a probabilistic classifier and then using the EM
algorithm \blue{(i)} to update (in an iterative, mutually recursive
way) the posterior probabilities that the classifier returns, and
\blue{(ii) to} re-estimate the class prevalence values of the test
set, until mutual consistency, defined as \blue{the situation in
which}
\begin{align}
  \label{eq:calib} 
  p_{U}(y) & \approx \sum_{\mathbf{x}\in U}\Pr(y|\mathbf{x})
\end{align}  
\noindent is achieved for all $y\in \mathcal{Y}$.

Quantification methods \textbf{SVM(KLD)}, \textbf{SVM(NKLD)},
\textbf{SVM(Q)}, belong instead to the ``structured output learning''
camp. Each of them is the result of instantiating the
SVM$_{\mathrm{perf}}$ structured output learner \cite{Joachims05} to
optimise a different loss function. SVM(KLD) \cite{Esuli:2015gh}
minimises the Kullback-Leibler Divergence (KLD); SVM(NKLD)
\cite{Esuli:2014uq} minimises a version of KLD normalised \blue{by
means of} the logistic function; SVM(Q) \cite{Barranquero:2015fr}
minimises \blue{Q,} the harmonic mean of a classification-oriented
loss (recall) and a quantification-oriented loss ($\rae$). Each of
these learners generates a ``quantification-oriented'' classifier, and
the quantification method consists of performing CC by using this
classifier.  These three learners inherently generate \textit{binary}
quantifiers (since SVM$_{\mathrm{perf}}$ is an algorithm for learning
binary predictors only), but we adapt them to work on single-label
multiclass quantification. This adaptation consists of training one
binary quantifier for each class in
$\mathcal{Y}=\{\oplus,\odot,\ominus\}$ by applying a one-vs-all
strategy. Once applied to a sample, these three binary quantifiers
produce a vector of three estimated prevalence values, one for each
class in $\mathcal{Y}=\{\oplus,\odot,\ominus\}$; we then L1-normalize
this vector so as to make the three class prevalence estimates sum up
to one (this is also the strategy followed in [GS2016]).


\subsection{Additional quantification methods}
\label{sec:additional}

\noindent From the ``structured output learning'' camp we also
consider \textbf{SVM(AE)} and \textbf{SVM(RAE)}, i.e., variants of the
\blue{above-mentioned methods} that minimise (instead of KLD, NKLD, or
Q) the AE and RAE measures, since these latter are, for reasons
discussed in Section~\ref{sec:measures}, the evaluation measures used
in this paper for evaluating the quantification accuracy of our
systems. We consider SVM(AE) only when using AE as the evaluation
measure, and we consider SVM(RAE) only when using RAE as the
evaluation measure; this obeys the principle that a sensible user,
after deciding the evaluation measure to use for their
experiments,\footnote{Quantification is a task in which deciding the
right evaluation measure to use for one's application is of critical
importance; in fact,~\cite{Sebastiani:2020qf} argues that some
applications demand measures such as AE, while the requirements of
other applications are best mirrored in measures such as RAE.}  would
instantiate SVM$_{\mathrm{perf}}$ with that measure, and not with
others. These methods have never been used before in the literature,
but are obvious variants of the last three methods we have described.

We also include two methods based on the notion of
\emph{quantification ensemble}
\cite{Perez-Gallego:2017wt,Perez-Gallego:2019vl}. Each such ensemble
consists of $n$ base quantifiers, trained \blue{from} randomly drawn
samples of \blue{$q$} documents each, \blue{where these samples are}
characterised by different class prevalence values. At testing time,
class prevalence values are estimated as the average of the estimates
returned by the base members of the ensemble. We include two
ensemble-based methods recently proposed
by Pérez-Gállego et al.~\cite{Perez-Gallego:2019vl}; in both methods, a selection of
members for inclusion in the final ensemble is performed before
computing the final estimate. The first method we consider is
\textbf{\EPACCPtr}, a method based on an ensemble of PACC-based
quantifiers to which a dynamic selection policy is applied. This
policy consists of selecting the $n/2$ base quantifiers that have been
trained \blue{on the $n/2$ samples} characterised by the prevalence
values most similar to the one being tested upon (where similarity was
previously estimated using all members in the ensemble). We further
consider \textbf{\EPACCAE}, a method which performs a static selection
of the $n/2$ members that deliver the smallest absolute error on the
training samples.
In our experiments
we use $n$=50 and
\blue{$q$}=\blue{1,000}.

We also report results for \textbf{HDy}
\cite{Gonzalez-Castro:2013fk}, a probabilistic binary quantification
method that views quantification as \blue{the} problem of minimising
the divergence (measured in terms of the Hellinger Distance) between
two cumulative distributions of posterior probabilities returned by
the classifier, one coming from the unlabelled examples and the other
coming from a validation set. HDy looks for the mixture parameter
$\alpha$ that best fits the validation distribution (consisting of a
mixture of a ``positive'' and a ``negative'' distribution) to the
unlabelled distribution, and returns $\alpha$ as the estimated
prevalence of the positive class. We adapt the model to the
single-label multiclass scenario by using the one-vs-all strategy as
described above for the methods based on SVM$_{\mathrm{perf}}$.

$\mathrm{ACC}$ and $\mathrm{PACC}$ define
two simple linear \emph{adjustments} to be applied to the aggregated
scores returned by general-purpose classifiers.  We also use a more
recently proposed adjustment method based on deep learning, called
\textbf{QuaNet} \cite{Esuli:2018rm}.  QuaNet models a neural
\emph{non-linear} adjustment by taking as input (i) all the class
prevalence values as estimated by CC, ACC, PCC, PACC, and SLD;
(ii) the posterior probabilities $\Pr(y|\mathbf{x})$ for each document
$\mathbf{x}$ and for each class $y\in\mathcal{Y}$, and (iii) embedded
representations of the documents. As the method for generating the
document embeddings we simply perform principal component analysis and
retain the 100 most informative components.\footnote{Note that, since
the datasets we use are available not in raw form but in vector form,
we cannot resort to common methods for generating document embeddings,
e.g., methods that use recurrent, convolutional, or transformer
architectures that directly process the raw text.}
QuaNet relies on a recurrent neural network (a bidirectional LSTM) to
produce ``sample embeddings'' (i.e., dense, multi-dimensional
representations of the test samples as observed from the input data),
which are then concatenated with the class prevalence estimates
obtained by CC, ACC, PCC, PACC, and SLD, and then used to generate the
final prevalence estimates by transforming this vector through a set
of feed-forward layers (of size \blue{1,024} and 512), followed by
ReLU activations and dropout (with drop probability set to 0.5).


\subsection{Underlying classifiers}
\label{sec:underlying}

\noindent Consistently with [GS2016], as the classifier underlying CC,
ACC, PCC, PACC, and SLD, we use one trained \blue{by means of}
L2-regularised logistic regression \blue{(LR)}; we also do the same
for E(PACC)$_{\mathrm{Ptr}}$, E(PACC)$_{\mathrm{AE}}$, HDy, and
QuaNet. The reasons of this choice are the same as described in
[GS2016], i.e., the fact that logistic regression is known to
\blue{deliver} very good classification accuracy across a variety of
application domains, and \blue{the fact} that a classifier trained
\blue{by means of} LR returns posterior probabilities that tend to be
fairly well-calibrated, a fact which is of fundamental importance for
methods such as PCC, PACC, SLD, HDy, and QuaNet. By using the same
learner used in [GS2016] we also allow a more direct comparison of
results.

As specified above, the classifier underlying SVM(KLD), SVM(NKLD),
SVM(Q), SVM(AE), SVM(RAE), is one trained \blue{by means of}
SVM$_{\mathrm{perf}}$.



\subsection{Datasets}
\label{sec:datasets}

\noindent The datasets on which we run our experiments are the same 11
datasets on which the experiments of [GS2016] were carried out, and
whose characteristics are described succinctly in
Table~\ref{tab:datasetscharacteristics}. As already noted at the end
of Section~\ref{sec:intro},~[GS2016] makes these datasets available
already in vector form; we refer to [GS2016] for a fuller description
of these datasets.

\blue{Note that [GS2016] had generated these vectors by using
state-of-the-art, tweet-specific preprocessing, which included, e.g.,
URL normalisation, detection of exclamation and/or question marks,
emoticon recognition, and computation of ``the number of all-caps
tokens, (...), the number of hashtags, the number of negated contexts,
the number of sequences of exclamation and/or question marks, and the
number of elongated words'' [GS2016, \S 4.1]; in other words, every
effort was made in [GS2016] to squeeze every little bit of information
from these tweets, in a tweet-specific way, in order to enhance
accuracy as much as possible.}

In the experiments described in this paper we perform feature
selection by discarding all features that occur in fewer than 5
training documents.

\begin{table}[tb]
  \caption{Datasets used in this work and their main characteristics.
  Columns $L_{\mathrm{Tr}}$, $L_{\mathrm{Va}}$, $U$ contain the
  numbers of tweets in the training set, held-out validation set, and
  test set, respectively. Column ``Shift'' contains the values of
  distribution shift between
  $L\equiv L_{\mathrm{Tr}}\bigcup L_{\mathrm{Va}}$ and $U$, measured
  in terms of absolute error\blue{, columns $p_{L}({\oplus})$,
  $p_{L}({\odot})$, and $p_{L}({\ominus})$ contain the class
  prevalence values of our three classes of interest in the training
  set $L$, while columns $p_{U}({\oplus})$, $p_{U}({\odot})$, and
  $p_{U}({\ominus})$ contain the class prevalence values for the
  unlabelled set $U$.}}
  \begin{center}
    \resizebox{\textwidth}{!}{
    \begin{tabular}{c|rrrrr|rrr|rrr}
      Dataset & \multicolumn{1}{c}{$L_{\mathrm{Tr}}$}  & \multicolumn{1}{c}{$L_{\mathrm{Va}}$} & \multicolumn{1}{c}{$U$} & \multicolumn{1}{c}{Total} & \multicolumn{1}{c|}{Shift}  & \blue{$p_{L}({\oplus})$} & \blue{$p_{L}({\odot})$} & \blue{$p_{L}({\ominus})$}
      & \blue{$p_{U}({\oplus})$} & \blue{$p_{U}({\odot})$} & \blue{$p_{U}({\ominus})$}\\
      \hline
      GASP             & 7,532 & 1,256 & 3,765 & 12,553 & 0.0094 & \blue{0.421} & \blue{ 0.496} & \blue{ 0.082} & \blue{0.407} & \blue{ 0.507} & \blue{ 0.086} \\ 
      HCR              &   797 & 797   & 798   &  2,392 & 0.0630 & \blue{0.546} & \blue{ 0.211} & \blue{ 0.243} & \blue{0.640} & \blue{ 0.167} & \blue{ 0.193} \\ 
      OMD              & 1,576 & 263   & 787   &  2,626 & 0.0171 & \blue{0.463} & \blue{ 0.271} & \blue{ 0.266} & \blue{0.437} & \blue{ 0.283} & \blue{ 0.280} \\ 
      Sanders     	   & 1,847 & 308   & 923   &  3,078 & 0.0020 & \blue{0.161} & \blue{ 0.691} & \blue{ 0.148} & \blue{0.164} & \blue{ 0.688} & \blue{ 0.148} \\ 
      SemEval2013      & 9,684 & 1,654 & 3,813 & 15,151 & 0.0270 & \blue{0.159} & \blue{ 0.470} & \blue{ 0.372} & \blue{0.158} & \blue{ 0.430} & \blue{ 0.412} \\ 
      SemEval2014      & 9,684 & 1,654 & 1,853 & 13,191 & 0.1055 & \blue{0.159} & \blue{ 0.470} & \blue{ 0.372} & \blue{0.109} & \blue{ 0.361} & \blue{ 0.530} \\ 
      SemEval2015      & 9,684 & 1,654 & 2,390 & 13,728 & 0.0417 & \blue{0.159} & \blue{ 0.470} & \blue{ 0.372} & \blue{0.153} & \blue{ 0.413} & \blue{ 0.434} \\ 
      SemEval2016      & 6,000 & 2,000 & 2,000 & 10,000 & 0.0070 & \blue{0.157} & \blue{ 0.351} & \blue{ 0.492} & \blue{0.163} & \blue{ 0.341} & \blue{ 0.497} \\ 
      SST              & 2,546 &   425 & 1,271 &  4,242 & 0.0357 & \blue{0.261} & \blue{ 0.452} & \blue{ 0.288} & \blue{0.207} & \blue{ 0.481} & \blue{ 0.312} \\ 
      WA  			   & 1,872 &   312 &   936 &  3,120 & 0.0208 & \blue{0.305} & \blue{ 0.414} & \blue{ 0.281} & \blue{0.282} & \blue{ 0.446} & \blue{ 0.272} \\ 
      WB   			   & 3,650 &   609 & 1,823 &  6,082 & 0.0023 & \blue{0.270} & \blue{ 0.392} & \blue{ 0.337} & \blue{0.274} & \blue{ 0.392} & \blue{ 0.335} \\ \hline
      Average          & 4,988 &   994 & 1,851 &  7,833 & 0.0301 & \blue{0.278} & \blue{0.426} & \blue{0.296} & \blue{0.272} & \blue{0.410} & \blue{0.318} \\ 
    \end{tabular}
    }
  \end{center}
  \label{tab:datasetscharacteristics}
\end{table}

According to the principles of the APP, as described in
Section~\ref{sec:appnpp}, for each of the 11 datasets we here extract
multiple samples from the test set, according to the following
protocol. For each different triple
$(p({\oplus}),p({\odot}),p({\ominus}))$ of class prevalence values
such that each class prevalence is in the finite
set $P=\{$0.00, 0.05, ..., 0.95, 1.00$\}$ and such that the three
values sum up to 1, we extract $m$ random samples of $q$ documents
each such that the extracted samples exhibit the class prevalence
values described by the triple.
In these experiments we use $m=25$ and $q=100$. For each label
$y\in\{\oplus,\odot,\ominus\}$ and for each sample, the extraction is
carried out \blue{by means of} sampling without
replacement.\footnote{Here it is possible to always use sampling
without replacement because each test set contains at least $q=100$
documents for each label $y\in\{\oplus,\odot,\ominus\}$. If a certain
test set contained fewer than $q=100$ documents for some label
$y\in\{\oplus,\odot,\ominus\}$, for that label and that test set it
would be necessary to use sampling with replacement.}

It is easy to verify that there exist $|P|(|P|+1)/2=231$ different
triples with values in $P$.\footnote{\label{foot:samples}This
\blue{follows} from the fact that, when $p({\oplus})(\sigma)=0.00$,
there exist 21 different pairs
$(p({\odot})(\sigma),p({\ominus})(\sigma))$ with values in $P$; when
$p({\oplus})(\sigma)=0.05$, there exist 20 different such pairs; ...;
and when $p({\oplus})(\sigma)=1.00$, there exists just 1 such
pair. The total number of combinations is thus
$\sum_{i=1}^{21}i=\frac{21\cdot22}{2}=231$.} Our experimentation of a
given quantification method $M$ on a given dataset thus consists of
training $M$ on the training tweets $L_{\mathrm{Tr}}$, using the
validation tweets $L_{\mathrm{Va}}$ for optimising the
hyperparameters, retraining $M$ on the entire labelled set
$L\equiv L_{\mathrm{Tr}}\bigcup L_{\mathrm{Va}}$ using the optimal
hyperparameter values, and testing the trained system on each of the
25$\times$231=\blue{5,775} samples extracted from the test set
$U$. This is sharply different from [GS20216], where the
experimentation of a quantification method $M$ on a given dataset
consists of testing the trained system on one sample only, i.e., on
the entire set $U$.


\subsection{Parameter optimisation}
\label{sec:paropt}

\noindent Parameter optimisation is an important factor, that could
bias, if not carried out properly, a comparative experimentation of
different quantification methods. As we have argued elsewhere
\cite{Moreo:2021sp}, when the quantification method is of the
aggregative type, for this experimentation to be unbiased, not only it
is important to optimise the hyperparameters of the classifier that
underlies the quantification method, but it is also important that
this optimisation is carried out using a quantification-oriented loss,
and not a classification-oriented \blue{loss}.

In order to optimise a quantification-oriented loss it is necessary to
test each hyperparameter setting on multiple samples extracted from
the held-out validation set, in the style of the evaluation described
in Section~\ref{sec:datasets}. In order to do this, for each
combination of class prevalence values we extract, from the held-out
validation set of each dataset, $m$ samples of $q$ documents each,
again using class prevalence values in $P=\{$0.00, 0.05, ..., 0.95,
1.00$\}$. Here we use $m=5$ and $q=100$; we use a value of $m$ five
times smaller than in the evaluation phase (see
Section~\ref{sec:datasets}) in order to keep the computational cost of
the parameter optimisation phase within acceptable bounds.

For each label $y\in\{\oplus,\odot,\ominus\}$ and for each sample, the
extraction is carried out by sampling without replacement if the test
set contains at least $p_{y}\cdot q$ examples, and by sampling with
replacement otherwise.\footnote{Unlike when extracting samples in the
evaluation phase (see Section~\ref{sec:datasets}), it is here
sometimes necessary to use sampling with replacement because, in some
dataset, the validation set does not contain at least 100 documents
per class.}

In the experiments \blue{that} we report in this paper, the
hyperparameter that we optimise is the $C$ \blue{hyperparameter} (that
determines the trade-off between the margin and the training error) of
both LR and SVM$_{\mathrm{perf}}$; for this we carry out a grid search
in the range $C \in \{10^{i}\}$, with $i\in [-4, -3,\ldots, +4,
+5]$. We optimise this parameter by using, as a loss function, either
the AE measure (the corresponding results are reported in
Table~\ref{tab:maeresults}) or the RAE measure
(Table~\ref{tab:mraeresults}). We evaluate the former batch of
experiments only in terms of AE and the latter batch only in terms of
RAE, following the principle that, once a user knew the measure to be
used in the evaluation, they would carry out the parameter
optimisation phase in terms of exactly that measure.

Hereafter, \blue{with} the notation $M^{D}$ we will indicate
quantification method $M$ with the parameters of the learner optimised
using measure $D$.


\section{Results}
\label{sec:results}

\noindent Table~\ref{tab:maeresults} reports AE results obtained by
the quantification methods of Sections~\ref{sec:methods} and
\ref{sec:additional} as tested on the 11 datasets of
Section~\ref{sec:datasets}, while Table~\ref{tab:mraeresults} does the
same for RAE. The tables also report the results of a paired sample,
two-tailed t-test that we have run, at different confidence levels, in
order to check if other methods are different or not, in a
statistically significant sense, from the best-performing one.

\begin{table}[tb]
  \caption{Values of AE obtained in our experiments; each value
  is the average across 5,775 values, each obtained on a different
  sample.  \textbf{Boldface} indicates the best method for a given
  dataset.  Superscripts $\dag$ and $\ddag$ denote the methods (if
  any) whose scores are \emph{not} statistically significantly
  different from the best one according to a paired sample, two-tailed
  t-test at different confidence levels: symbol $\dag$ indicates \blue{that}
  $0.001<p$-value $<0.05$ while symbol $\ddag$ indicates \blue{that}
  $0.05\leq p$-value. The absence of any such symbol indicates \blue{that}
  $p$-value $\leq 0.001$ (i.e., that the performance of the method is
  statistically significantly different from that of the best method).
  For ease of readability, for each dataset we colour-code cells in
  intense green for the best result, intense red for the worst result,
  and an interpolated tone for the scores in-between.
  }
  \begin{center}
    
        \resizebox{\textwidth}{!}{%
                \begin{tabular}{|c||c|c|c|c|c|c|c|c||c|c|c|c|c|} \hline
                  & \multicolumn{8}{c||}{Methods tested in [GS2016]} & 
                    \multicolumn{5}{c|}{Newly added methods} \\ \hline
                 & \side{CC$^{\mathrm{AE}}$} & \side{ACC$^{\mathrm{AE}}$} & \side{PCC$^{\mathrm{AE}}$} & \side{PACC$^{\mathrm{AE}}$} & \side{SLD$^{\mathrm{AE}}$} & \side{SVM(Q)$^{\mathrm{AE}}$} & \side{SVM(KLD)$^{\mathrm{AE}}$} & \side{SVM(NKLD)$^{\mathrm{AE}}$} & \side{SVM(AE)$^{\mathrm{AE}}$} & \side{E(PACC)$_\mathrm{Ptr}^{\mathrm{AE}}$} & \side{E(PACC)$_\mathrm{AE}^{\mathrm{AE}}$} & \side{HDy$^{\mathrm{AE}}$} & \side{QuaNet$^{\mathrm{AE}}$} \\\hline
GASP &  0.093\pdag\cellcolor{red!3} &  0.052\pdag\cellcolor{green!40} &  0.124\pdag\cellcolor{red!36} &  0.044\pdag\cellcolor{green!48} & \textbf{0.043}\pdag\cellcolor{green!50} &  0.119\pdag\cellcolor{red!31} &  0.114\pdag\cellcolor{red!26} &  0.110\pdag\cellcolor{red!21} &  0.136\pdag\cellcolor{red!50} &  0.065\pdag\cellcolor{green!25} &  0.049\pdag\cellcolor{green!43} &  0.086\pdag\cellcolor{green!3} &  0.046\pdag\cellcolor{green!46} \\\hline
HCR &  0.130\pdag\cellcolor{red!18} &  0.102\pdag\cellcolor{green!12} &  0.158\pdag\cellcolor{red!50} &  0.074\pdag\cellcolor{green!45} &  0.078\pdag\cellcolor{green!40} &  0.150\pdag\cellcolor{red!40} &  0.143\pdag\cellcolor{red!33} &  0.138\pdag\cellcolor{red!27} &  0.158\pdag\cellcolor{red!49} &  0.084\pdag\cellcolor{green!33} &  0.071$^{\dag}$\cellcolor{green!48} & \textbf{0.070}\pdag\cellcolor{green!50} &  0.099\pdag\cellcolor{green!17} \\\hline
OMD &  0.114\pdag\cellcolor{red!17} &  0.086\pdag\cellcolor{green!14} &  0.126\pdag\cellcolor{red!32} &  0.067\pdag\cellcolor{green!37} & \textbf{0.055}\pdag\cellcolor{green!50} &  0.141\pdag\cellcolor{red!50} &  0.124\pdag\cellcolor{red!30} &  0.139\pdag\cellcolor{red!47} &  0.116\pdag\cellcolor{red!20} &  0.084\pdag\cellcolor{green!16} &  0.075\pdag\cellcolor{green!27} &  0.119\pdag\cellcolor{red!24} &  0.087\pdag\cellcolor{green!12} \\\hline
Sanders &  0.114\pdag\cellcolor{red!12} &  0.058\pdag\cellcolor{green!38} &  0.138\pdag\cellcolor{red!33} &  0.049\pdag\cellcolor{green!46} & \textbf{0.045}\pdag\cellcolor{green!50} &  0.140\pdag\cellcolor{red!35} &  0.141\pdag\cellcolor{red!35} &  0.110\pdag\cellcolor{red!8} &  0.157\pdag\cellcolor{red!50} &  0.076\pdag\cellcolor{green!22} &  0.058\pdag\cellcolor{green!37} &  0.087\pdag\cellcolor{green!12} &  0.079\pdag\cellcolor{green!19} \\\hline
SemEval13 &  0.115\pdag\cellcolor{red!5} &  0.086\pdag\cellcolor{green!37} &  0.143\pdag\cellcolor{red!47} & \textbf{0.078}\pdag\cellcolor{green!50} &  0.097\pdag\cellcolor{green!20} &  0.129\pdag\cellcolor{red!26} &  0.144\pdag\cellcolor{red!50} &  0.134\pdag\cellcolor{red!34} &  0.143\pdag\cellcolor{red!49} &  0.102\pdag\cellcolor{green!12} &  0.093\pdag\cellcolor{green!26} &  0.114\pdag\cellcolor{red!4} &  0.078$^{\ddag}$\cellcolor{green!49} \\\hline
SemEval14 &  0.105\pdag\cellcolor{red!11} &  0.060\pdag\cellcolor{green!43} &  0.136\pdag\cellcolor{red!50} & \textbf{0.054}\pdag\cellcolor{green!50} &  0.076\pdag\cellcolor{green!23} &  0.127\pdag\cellcolor{red!38} &  0.128\pdag\cellcolor{red!40} &  0.122\pdag\cellcolor{red!32} &  0.134\pdag\cellcolor{red!47} &  0.096\pdag\cellcolor{red!1} &  0.067\pdag\cellcolor{green!34} &  0.083\pdag\cellcolor{green!15} &  0.059\pdag\cellcolor{green!44} \\\hline
SemEval15 &  0.128\pdag\cellcolor{red!7} &  0.103\pdag\cellcolor{green!40} &  0.148\pdag\cellcolor{red!47} &  0.101$^{\dag}$\cellcolor{green!44} &  0.104\pdag\cellcolor{green!39} &  0.143\pdag\cellcolor{red!36} &  0.150\pdag\cellcolor{red!50} &  0.145\pdag\cellcolor{red!40} &  0.144\pdag\cellcolor{red!39} &  0.114\pdag\cellcolor{green!19} &  0.112\pdag\cellcolor{green!22} &  0.105\pdag\cellcolor{green!37} & \textbf{0.098}\pdag\cellcolor{green!50} \\\hline
SemEval16 &  0.146\pdag\cellcolor{red!8} &  0.147\pdag\cellcolor{red!9} &  0.171\pdag\cellcolor{red!40} &  0.118\pdag\cellcolor{green!28} & \textbf{0.102}\pdag\cellcolor{green!50} &  0.167\pdag\cellcolor{red!35} &  0.154\pdag\cellcolor{red!18} &  0.165\pdag\cellcolor{red!32} &  0.178\pdag\cellcolor{red!50} &  0.131\pdag\cellcolor{green!11} &  0.132\pdag\cellcolor{green!9} &  0.167\pdag\cellcolor{red!35} &  0.103$^{\ddag}$\cellcolor{green!47} \\\hline
SST &  0.110\pdag\cellcolor{red!15} &  0.083\pdag\cellcolor{green!15} &  0.140\pdag\cellcolor{red!50} &  0.057\pdag\cellcolor{green!45} & \textbf{0.054}\pdag\cellcolor{green!50} &  0.136\pdag\cellcolor{red!46} &  0.113\pdag\cellcolor{red!19} &  0.128\pdag\cellcolor{red!36} &  0.126\pdag\cellcolor{red!33} &  0.063\pdag\cellcolor{green!38} &  0.054$^{\ddag}$\cellcolor{green!49} &  0.097\pdag\cellcolor{red!0} &  0.069\pdag\cellcolor{green!31} \\\hline
WA &  0.082\pdag\cellcolor{red!10} &  0.056\pdag\cellcolor{green!24} &  0.082\pdag\cellcolor{red!10} &  0.043\pdag\cellcolor{green!41} & \textbf{0.037}\pdag\cellcolor{green!50} &  0.111\pdag\cellcolor{red!50} &  0.100\pdag\cellcolor{red!34} &  0.063\pdag\cellcolor{green!14} &  0.071\pdag\cellcolor{green!4} &  0.043\pdag\cellcolor{green!42} &  0.041\pdag\cellcolor{green!44} &  0.043\pdag\cellcolor{green!42} &  0.053\pdag\cellcolor{green!28} \\\hline
WB &  0.077\pdag\cellcolor{red!11} &  0.043\pdag\cellcolor{green!35} &  0.083\pdag\cellcolor{red!19} &  0.035\pdag\cellcolor{green!46} & \textbf{0.032}\pdag\cellcolor{green!50} &  0.106\pdag\cellcolor{red!50} &  0.084\pdag\cellcolor{red!19} &  0.103\pdag\cellcolor{red!46} &  0.069\pdag\cellcolor{red!0} &  0.048\pdag\cellcolor{green!28} &  0.041\pdag\cellcolor{green!37} &  0.044\pdag\cellcolor{green!33} &  0.046\pdag\cellcolor{green!31} \\\hline
\hline
Average &  0.110\pdag\cellcolor{red!16} &  0.080$^{\ddag}$\cellcolor{green!28} &  0.132\pdag\cellcolor{red!47} & \textbf{0.065}\pdag\cellcolor{green!50} &  0.066$^{\ddag}$\cellcolor{green!49} &  0.134\pdag\cellcolor{red!50} &  0.127\pdag\cellcolor{red!40} &  0.123\pdag\cellcolor{red!35} &  0.130\pdag\cellcolor{red!45} &  0.082$^{\ddag}$\cellcolor{green!25} &  0.072$^{\ddag}$\cellcolor{green!40} &  0.092$^{\dag}$\cellcolor{green!10} &  0.074$^{\ddag}$\cellcolor{green!36} \\\hline

            \end{tabular}%
            }
        
  \end{center}
  \label{tab:maeresults}
\end{table}
    
\begin{table}[tb]
  \caption{\blue{Values of RAE obtained in our experiments; each value
  is the average across 5,775 values, each obtained on a different
  sample.  \textbf{Boldface} indicates the best method for a given
  dataset.  Superscripts $\dag$ and $\ddag$ denote the methods (if
  any) whose scores are \emph{not} statistically significantly
  different from the best one according to a paired sample, two-tailed
  t-test at different confidence levels: symbol $\dag$ indicates
  \blue{that} $0.001<p$-value $<0.05$ while symbol $\ddag$ indicates
  \blue{that} $0.05\leq p$-value. The absence of any such symbol
  indicates \blue{that} $p$-value $\leq 0.001$ (i.e., that the
  performance of the method is statistically significantly different
  from that of the best method).  For ease of readability, for each
  dataset we colour-code cells in intense green for the best result,
  intense red for the worst result, and an interpolated tone for the
  scores in-between.}}
  \begin{center}
    
        \resizebox{\textwidth}{!}{%
                \begin{tabular}{|c||c|c|c|c|c|c|c|c||c|c|c|c|c|} \hline
                  & \multicolumn{8}{c||}{Methods tested in [GS2016]} & 
                    \multicolumn{5}{c|}{Newly added methods} \\ \hline
                 & \side{CC$^{\mathrm{RAE}}$} & \side{ACC$^{\mathrm{RAE}}$} & \side{PCC$^{\mathrm{RAE}}$} & \side{PACC$^{\mathrm{RAE}}$} & \side{SLD$^{\mathrm{RAE}}$} & \side{SVM(Q)$^{\mathrm{RAE}}$} & \side{SVM(KLD)$^{\mathrm{RAE}}$} & \side{SVM(NKLD)$^{\mathrm{RAE}}$} & \side{SVM(RAE)$^{\mathrm{RAE}}$} & \side{E(PACC)$_\mathrm{Ptr}^{\mathrm{RAE}}$} & \side{E(PACC)$_\mathrm{RAE}^{\mathrm{RAE}}$} & \side{HDy$^{\mathrm{RAE}}$} & \side{QuaNet$^{\mathrm{RAE}}$} \\\hline
GASP &  2.850\pdag\cellcolor{red!13} &  0.512\pdag\cellcolor{green!45} &  3.490\pdag\cellcolor{red!30} &  0.722\pdag\cellcolor{green!40} & \textbf{0.337}\pdag\cellcolor{green!50} &  3.835\pdag\cellcolor{red!38} &  3.260\pdag\cellcolor{red!24} &  3.461\pdag\cellcolor{red!29} &  3.411\pdag\cellcolor{red!28} &  2.361\pdag\cellcolor{red!1} &  1.402\pdag\cellcolor{green!22} &  0.644\pdag\cellcolor{green!42} &  4.270\pdag\cellcolor{red!50} \\\hline
HCR &  3.982\pdag\cellcolor{red!28} &  1.942\pdag\cellcolor{green!16} &  4.151\pdag\cellcolor{red!32} &  1.332\pdag\cellcolor{green!30} & \textbf{0.454}\pdag\cellcolor{green!50} &  4.939\pdag\cellcolor{red!50} &  4.236\pdag\cellcolor{red!34} &  4.197\pdag\cellcolor{red!33} &  4.041\pdag\cellcolor{red!29} &  2.169\pdag\cellcolor{green!11} &  1.990\pdag\cellcolor{green!15} &  0.517\pdag\cellcolor{green!48} &  4.214\pdag\cellcolor{red!33} \\\hline
OMD &  3.495\pdag\cellcolor{red!23} &  0.884\pdag\cellcolor{green!39} &  3.776\pdag\cellcolor{red!30} &  0.552\pdag\cellcolor{green!47} & \textbf{0.469}\pdag\cellcolor{green!50} &  4.578\pdag\cellcolor{red!50} &  3.844\pdag\cellcolor{red!32} &  4.481\pdag\cellcolor{red!47} &  3.295\pdag\cellcolor{red!18} &  2.479\pdag\cellcolor{green!1} &  1.840\pdag\cellcolor{green!16} &  0.881\pdag\cellcolor{green!39} &  2.296\pdag\cellcolor{green!5} \\\hline
Sanders &  3.296\pdag\cellcolor{red!22} &  0.791\pdag\cellcolor{green!40} &  3.687\pdag\cellcolor{red!32} &  0.990\pdag\cellcolor{green!35} & \textbf{0.432}\pdag\cellcolor{green!50} &  4.377\pdag\cellcolor{red!50} &  3.596\pdag\cellcolor{red!30} &  3.533\pdag\cellcolor{red!28} &  3.767\pdag\cellcolor{red!34} &  2.342\pdag\cellcolor{green!1} &  1.559\pdag\cellcolor{green!21} &  0.504\pdag\cellcolor{green!48} &  1.943\pdag\cellcolor{green!11} \\\hline
SemEval13 &  3.117\pdag\cellcolor{red!24} &  1.469\pdag\cellcolor{green!22} &  3.720\pdag\cellcolor{red!42} &  1.244\pdag\cellcolor{green!28} & \textbf{0.491}\pdag\cellcolor{green!50} &  3.998\pdag\cellcolor{red!50} &  3.743\pdag\cellcolor{red!42} &  3.960\pdag\cellcolor{red!48} &  3.588\pdag\cellcolor{red!38} &  2.162\pdag\cellcolor{green!2} &  1.602\pdag\cellcolor{green!18} &  1.027\pdag\cellcolor{green!34} &  1.712\pdag\cellcolor{green!15} \\\hline
SemEval14 &  3.079\pdag\cellcolor{red!24} &  1.414\pdag\cellcolor{green!20} &  3.699\pdag\cellcolor{red!41} &  1.271\pdag\cellcolor{green!24} & \textbf{0.325}\pdag\cellcolor{green!50} &  4.018\pdag\cellcolor{red!50} &  3.535\pdag\cellcolor{red!36} &  3.741\pdag\cellcolor{red!42} &  3.364\pdag\cellcolor{red!32} &  2.261\pdag\cellcolor{red!2} &  1.691\pdag\cellcolor{green!13} &  0.523\pdag\cellcolor{green!44} &  1.384\pdag\cellcolor{green!21} \\\hline
SemEval15 &  3.608\pdag\cellcolor{red!27} &  1.695\pdag\cellcolor{green!27} &  4.022\pdag\cellcolor{red!38} &  1.884\pdag\cellcolor{green!21} & \textbf{0.889}\pdag\cellcolor{green!50} &  4.417\pdag\cellcolor{red!50} &  4.031\pdag\cellcolor{red!39} &  4.264\pdag\cellcolor{red!45} &  3.847\pdag\cellcolor{red!33} &  2.552\pdag\cellcolor{green!2} &  2.233\pdag\cellcolor{green!11} &  1.275\pdag\cellcolor{green!39} &  1.866\pdag\cellcolor{green!22} \\\hline
SemEval16 &  4.594\pdag\cellcolor{red!30} &  2.994\pdag\cellcolor{green!7} &  5.191\pdag\cellcolor{red!44} &  2.815\pdag\cellcolor{green!12} & \textbf{1.216}\pdag\cellcolor{green!50} &  5.430\pdag\cellcolor{red!50} &  4.880\pdag\cellcolor{red!36} &  5.278\pdag\cellcolor{red!46} &  4.988\pdag\cellcolor{red!39} &  4.090\pdag\cellcolor{red!18} &  4.057\pdag\cellcolor{red!17} &  1.773\pdag\cellcolor{green!36} &  4.026\pdag\cellcolor{red!16} \\\hline
SST &  4.207\pdag\cellcolor{red!43} &  0.972\pdag\cellcolor{green!37} &  4.226\pdag\cellcolor{red!44} &  1.042\pdag\cellcolor{green!35} &  0.534$^{\dag}$\cellcolor{green!48} &  4.446\pdag\cellcolor{red!50} &  3.621\pdag\cellcolor{red!29} &  3.535\pdag\cellcolor{red!26} &  3.804\pdag\cellcolor{red!33} &  1.880\pdag\cellcolor{green!14} &  1.343\pdag\cellcolor{green!28} & \textbf{0.487}\pdag\cellcolor{green!50} &  1.783\pdag\cellcolor{green!17} \\\hline
WA &  2.493\pdag\cellcolor{red!18} &  0.540\pdag\cellcolor{green!42} &  2.706\pdag\cellcolor{red!25} &  0.512\pdag\cellcolor{green!43} & \textbf{0.313}\pdag\cellcolor{green!50} &  3.503\pdag\cellcolor{red!50} &  2.814\pdag\cellcolor{red!28} &  1.431\pdag\cellcolor{green!14} &  1.739\pdag\cellcolor{green!5} &  0.948\pdag\cellcolor{green!30} &  0.825\pdag\cellcolor{green!33} &  0.587\pdag\cellcolor{green!41} &  1.280\pdag\cellcolor{green!19} \\\hline
WB &  2.419\pdag\cellcolor{red!18} &  0.693\pdag\cellcolor{green!35} &  2.560\pdag\cellcolor{red!22} &  0.669\pdag\cellcolor{green!36} & \textbf{0.233}\pdag\cellcolor{green!50} &  3.440\pdag\cellcolor{red!50} &  2.525\pdag\cellcolor{red!21} &  2.243\pdag\cellcolor{red!12} &  1.926\pdag\cellcolor{red!2} &  0.975\pdag\cellcolor{green!26} &  0.790\pdag\cellcolor{green!32} &  0.283\pdag\cellcolor{green!48} &  1.205\pdag\cellcolor{green!19} \\\hline
\hline
Average &  3.376\pdag\cellcolor{red!26} &  1.264\pdag\cellcolor{green!30} &  3.748\pdag\cellcolor{red!36} &  1.185\pdag\cellcolor{green!32} & \textbf{0.518}\pdag\cellcolor{green!50} &  4.271\pdag\cellcolor{red!50} &  3.644\pdag\cellcolor{red!33} &  3.648\pdag\cellcolor{red!33} &  3.434\pdag\cellcolor{red!27} &  2.202\pdag\cellcolor{green!5} &  1.757\pdag\cellcolor{green!16} &  0.773$^{\ddag}$\cellcolor{green!43} &  2.362\pdag\cellcolor{green!0} \\\hline

            \end{tabular}%
            }
        
  \end{center}
  \label{tab:mraeresults}
\end{table}

An important aspect that emerges from \blue{these tables} is that the
behaviour of the different quantifiers is fairly consistent across our
11 datasets; in other words, when a method is a good performer on one
dataset, it tends to be a good performer \emph{on all
datasets}. Together with the fact that we test on a large set of
samples, and that these are characterised by values of distribution
shift across the entire range of all possible such shifts, this allows
us to be fairly confident in the conclusions that we draw from these
results.

A second observation is that three methods (ACC, PACC, and SLD) stand
out, since they perform consistently well across all datasets and for
both evaluation measures. In particular, SLD is the best method for
\blue{7 out of 11 datasets} (and is not different, in
a statistically significant sense, from the best method on yet another
dataset) when testing with AE, and for all 11 datasets when testing
with RAE. PACC also \blue{performs} very well, and is the best
performer for 3 out of 11 datasets when testing with AE. The fact that
both ACC and PACC tend to perform well shows that the intuition
according to which CC predictions should be ``\blue{adjusted}'' by
estimating the disposition of the classifier to assign class $y_{i}$
when class $y_{j}$ is the \blue{true label}, is valuable and robust to
varying levels of distribution shift. The same goes for SLD, although
SLD ``\blue{adjusts}'' the CC predictions differently, i.e., by
enforcing the mutual consistency (described by
Equation~\ref{eq:calib}) between the posterior probabilities and the
class prevalence estimates.

By contrast, these results show a generally disappointing performance
on the part of all methods based on structured output learning, i.e.,
on the SVM$_{\mathrm{perf}}$ learner. Note that the fact that
SVM(KLD), SVM(NKLD), SVM(Q) optimise a performance measure different
from the one used in the evaluation (AE or RAE) cannot be the cause of
this suboptimal performance, since this latter also characterises
SVM(AE) when tested with AE as the evaluation measure, and SVM(RAE)
when tested with RAE.

CC and PCC do no perform well either. If this was somehow to be
expected for CC, this is surprising for PCC, which always performs
worse than CC in our experiments, on all datasets and for both
performance measures. It would be tempting to conjecture that this
might be due to a supposedly insufficient quality of the posterior
probabilities returned by the underlying classifier; however, this
conjecture is implausible, since the quality of the posterior
probabilities did not prevent SLD from displaying sterling
performance, and PACC from \blue{performing} very well.

Contrary to the observations reported in
\cite{Perez-Gallego:2019vl}, the \EPACCPtr\ and \EPACCAE\ ensemble
methods fail to improve over the base quantifier (PACC) upon which
they are built. The likely reason for this discrepancy is that, while
Pérez-Gállego et al.~\cite{Perez-Gallego:2019vl} trained the base quantifiers on training
samples of the same size as the original training set (i.e., they use
$q=|L|$), we use smaller training samples (i.e., we use
$q$=\blue{1,000}) in order to keep training times within reasonable
bounds (this is also due to the fact that the datasets we consider in
this study are much larger than those used in
\cite{Perez-Gallego:2019vl}, not only in terms of the number of
instances but especially in terms of the number of
features).\footnote{For instance, our datasets always have a number of
features in the tens or hundreds of thousands, while in their case
this number if between 3 and 256.}

We now turn to comparing the results of our experiments with the ones
reported in [GS2016]. For doing this, for each dataset we rank, in
terms of their performance, the 8 quantification methods used in both
batches of experiments, and compare the rank positions obtained by
each method in the two
batches.\footnote{\blue{\label{foot:comparison}We only perform a
qualitative comparison (i.e., comparing ranks) and not a quantitative
one (i.e., comparing the obtained scores) because we think that this
latter would be misleading. The reason is that the evaluation carried
out in the [GS2016] paper and the one carried out here were run on
different data. For example, on dataset GASP and using AE as the
evaluation measure, SVM(KLD) obtains 0.017 in [GS2016] and 0.114 in
this paper, but these results are not comparable, since the above
figures are (i) the result of testing on just 1 sample (the unlabelled
set) in [GS2016], and (ii) the result of averaging across the results
obtained on the 5,775 samples (extracted from the unlabelled set)
described in Section~\ref{sec:datasets} in this paper. In general, for
the same dataset and evaluation measure, the results reported in this
paper are far worse than the ones reported in [GS2016], because the
experimental protocol adopted in this paper is far more challenging
than the one used in [GS2016] (since it involves testing on samples
whose distribution is very different from the distribution of the
training set).}}

The results of this comparison are reported in
Table~\ref{tab:maeranks} (for AE) and Table~\ref{tab:mraeranks} (for
RAE).
\begin{table}[tb]
  \caption{Rank positions of the quantification methods in our AE
  experiments, and (between parentheses) the rank positions obtained
  by the same methods in the evaluation of [GS2016]. \textbf{Boldface}
  indicates the best method in terms of average rank in our APP-based
  experiments, while \underline{underline} is used to indicate the
  same for the NPP-based experiments of [GS2016].}
  \begin{center}
    
        \resizebox{\textwidth}{!}{%
        \begin{tabular}{|c||c|c|c|c|c|c|c|c|} \hline
              & \multicolumn{8}{c|}{Methods tested in [GS2016]}  \\ \hline
         & \side{CC$^{\mathrm{AE}}$} & \side{ACC$^{\mathrm{AE}}$} & \side{PCC$^{\mathrm{AE}}$} & \side{PACC$^{\mathrm{AE}}$} & \side{SLD$^{\mathrm{AE}}$} & \side{SVM(Q)$^{\mathrm{AE}}$} & \side{SVM(KLD)$^{\mathrm{AE}}$} & \side{SVM(NKLD)$^{\mathrm{AE}}$}\\\hline
GASP  & 4 (5) \cellcolor{green!7} & 3 (3) \cellcolor{green!21} & 8 (2) \cellcolor{red!50} & 2 (1) \cellcolor{green!35} & 1 (6) \cellcolor{green!50} & 7 (8) \cellcolor{red!35} & 6 (4) \cellcolor{red!21} & 5 (7) \cellcolor{red!7}\\\hline
HCR  & 4 (5) \cellcolor{green!7} & 3 (2) \cellcolor{green!21} & 8 (1) \cellcolor{red!50} & 1 (3) \cellcolor{green!50} & 2 (7) \cellcolor{green!35} & 7 (8) \cellcolor{red!35} & 6 (4) \cellcolor{red!21} & 5 (6) \cellcolor{red!7}\\\hline
OMD  & 4 (6) \cellcolor{green!7} & 3 (3) \cellcolor{green!21} & 6 (1) \cellcolor{red!21} & 2 (2) \cellcolor{green!35} & 1 (8) \cellcolor{green!50} & 8 (7) \cellcolor{red!50} & 5 (4) \cellcolor{red!7} & 7 (5) \cellcolor{red!35}\\\hline
Sanders  & 5 (5) \cellcolor{red!7} & 3 (4) \cellcolor{green!21} & 6 (2) \cellcolor{red!21} & 2 (3) \cellcolor{green!35} & 1 (6) \cellcolor{green!50} & 7 (8) \cellcolor{red!35} & 8 (1) \cellcolor{red!50} & 4 (7) \cellcolor{green!7}\\\hline
SemEval13  & 4 (7) \cellcolor{green!7} & 2 (5) \cellcolor{green!35} & 7 (1) \cellcolor{red!35} & 1 (6) \cellcolor{green!50} & 3 (8) \cellcolor{green!21} & 5 (4) \cellcolor{red!7} & 8 (3) \cellcolor{red!50} & 6 (2) \cellcolor{red!21}\\\hline
SemEval14  & 4 (8) \cellcolor{green!7} & 2 (2) \cellcolor{green!35} & 8 (6) \cellcolor{red!50} & 1 (1) \cellcolor{green!50} & 3 (3) \cellcolor{green!21} & 6 (7) \cellcolor{red!21} & 7 (5) \cellcolor{red!35} & 5 (4) \cellcolor{red!7}\\\hline
SemEval15  & 4 (4) \cellcolor{green!7} & 2 (3) \cellcolor{green!35} & 7 (1) \cellcolor{red!35} & 1 (2) \cellcolor{green!50} & 3 (7) \cellcolor{green!21} & 5 (8) \cellcolor{red!7} & 8 (6) \cellcolor{red!50} & 6 (5) \cellcolor{red!21}\\\hline
SemEval16  & 3 (3) \cellcolor{green!21} & 4 (4) \cellcolor{green!7} & 8 (1) \cellcolor{red!50} & 2 (7) \cellcolor{green!35} & 1 (5) \cellcolor{green!50} & 7 (8) \cellcolor{red!35} & 5 (2) \cellcolor{red!7} & 6 (6) \cellcolor{red!21}\\\hline
SST  & 4 (2) \cellcolor{green!7} & 3 (5) \cellcolor{green!21} & 8 (1) \cellcolor{red!50} & 2 (8) \cellcolor{green!35} & 1 (3) \cellcolor{green!50} & 7 (6) \cellcolor{red!35} & 5 (4) \cellcolor{red!7} & 6 (7) \cellcolor{red!21}\\\hline
WA  & 5 (6) \cellcolor{red!7} & 3 (5) \cellcolor{green!21} & 6 (2) \cellcolor{red!21} & 2 (1) \cellcolor{green!35} & 1 (3) \cellcolor{green!50} & 8 (8) \cellcolor{red!50} & 7 (7) \cellcolor{red!35} & 4 (4) \cellcolor{green!7}\\\hline
WB  & 4 (2) \cellcolor{green!7} & 3 (4) \cellcolor{green!21} & 5 (1) \cellcolor{red!7} & 2 (3) \cellcolor{green!35} & 1 (5) \cellcolor{green!50} & 8 (6) \cellcolor{red!50} & 6 (8) \cellcolor{red!21} & 7 (7) \cellcolor{red!35}\\\hline
\hline
Average  & 4.1 (4.8) \cellcolor{green!4} & 2.8 (3.6) \cellcolor{green!27} & 7.0 (\underline{1.7}) \cellcolor{red!50} & \textbf{1.6} (3.4) \cellcolor{green!50} & \textbf{1.6} (5.5) \cellcolor{green!50} & 6.8 (7.1) \cellcolor{red!46} & 6.5 (4.4) \cellcolor{red!39} & 5.5 (5.5) \cellcolor{red!22}\\\hline

        \end{tabular}%
        }
        
  \end{center}
  \label{tab:maeranks}
\end{table}
\begin{table}[tb]
  \caption{\blue{Rank positions of the quantification methods in our
  RAE experiments, and (between parentheses) the rank positions
  obtained by the same methods in the evaluation of
  [GS2016]. \textbf{Boldface} indicates the best method in terms of
  average rank in our APP-based experiments, while
  \underline{underline} is used to indicate the same for the NPP-based
  experiments of [GS2016].}  }
  \begin{center}
    
        \resizebox{\textwidth}{!}{%
        \begin{tabular}{|c||c|c|c|c|c|c|c|c|} \hline
              & \multicolumn{8}{c|}{Methods tested in [GS2016]}  \\ \hline
         & \side{CC$^{\mathrm{RAE}}$} & \side{ACC$^{\mathrm{RAE}}$} & \side{PCC$^{\mathrm{RAE}}$} & \side{PACC$^{\mathrm{RAE}}$} & \side{SLD$^{\mathrm{RAE}}$} & \side{SVM(Q)$^{\mathrm{RAE}}$} & \side{SVM(KLD)$^{\mathrm{RAE}}$} & \side{SVM(NKLD)$^{\mathrm{RAE}}$}\\\hline
GASP  & 4 (5) \cellcolor{green!7} & 2 (4) \cellcolor{green!35} & 7 (3) \cellcolor{red!35} & 3 (2) \cellcolor{green!21} & 1 (6) \cellcolor{green!50} & 8 (8) \cellcolor{red!50} & 5 (1) \cellcolor{red!7} & 6 (7) \cellcolor{red!21}\\\hline
HCR  & 4 (4) \cellcolor{green!7} & 3 (2) \cellcolor{green!21} & 5 (1) \cellcolor{red!7} & 2 (3) \cellcolor{green!35} & 1 (7) \cellcolor{green!50} & 8 (8) \cellcolor{red!50} & 7 (5) \cellcolor{red!35} & 6 (6) \cellcolor{red!21}\\\hline
OMD  & 4 (6) \cellcolor{green!7} & 3 (3) \cellcolor{green!21} & 5 (1) \cellcolor{red!7} & 2 (2) \cellcolor{green!35} & 1 (8) \cellcolor{green!50} & 8 (7) \cellcolor{red!50} & 6 (4) \cellcolor{red!21} & 7 (5) \cellcolor{red!35}\\\hline
Sanders  & 4 (5) \cellcolor{green!7} & 2 (4) \cellcolor{green!35} & 7 (1) \cellcolor{red!35} & 3 (3) \cellcolor{green!21} & 1 (6) \cellcolor{green!50} & 8 (8) \cellcolor{red!50} & 6 (2) \cellcolor{red!21} & 5 (7) \cellcolor{red!7}\\\hline
SemEval13  & 4 (7) \cellcolor{green!7} & 3 (3) \cellcolor{green!21} & 5 (1) \cellcolor{red!7} & 2 (4) \cellcolor{green!35} & 1 (8) \cellcolor{green!50} & 8 (6) \cellcolor{red!50} & 6 (2) \cellcolor{red!21} & 7 (5) \cellcolor{red!35}\\\hline
SemEval14  & 4 (4) \cellcolor{green!7} & 3 (2) \cellcolor{green!21} & 6 (8) \cellcolor{red!21} & 2 (3) \cellcolor{green!35} & 1 (6) \cellcolor{green!50} & 8 (7) \cellcolor{red!50} & 5 (1) \cellcolor{red!7} & 7 (5) \cellcolor{red!35}\\\hline
SemEval15  & 4 (3) \cellcolor{green!7} & 2 (5) \cellcolor{green!35} & 5 (1) \cellcolor{red!7} & 3 (2) \cellcolor{green!21} & 1 (4) \cellcolor{green!50} & 8 (8) \cellcolor{red!50} & 6 (6) \cellcolor{red!21} & 7 (7) \cellcolor{red!35}\\\hline
SemEval16  & 4 (3) \cellcolor{green!7} & 3 (5) \cellcolor{green!21} & 6 (2) \cellcolor{red!21} & 2 (7) \cellcolor{green!35} & 1 (4) \cellcolor{green!50} & 8 (8) \cellcolor{red!50} & 5 (1) \cellcolor{red!7} & 7 (6) \cellcolor{red!35}\\\hline
SST  & 6 (3) \cellcolor{red!21} & 2 (5) \cellcolor{green!35} & 7 (1) \cellcolor{red!35} & 3 (6) \cellcolor{green!21} & 1 (2) \cellcolor{green!50} & 8 (7) \cellcolor{red!50} & 5 (4) \cellcolor{red!7} & 4 (8) \cellcolor{green!7}\\\hline
WA  & 5 (5) \cellcolor{red!7} & 3 (4) \cellcolor{green!21} & 6 (2) \cellcolor{red!21} & 2 (1) \cellcolor{green!35} & 1 (3) \cellcolor{green!50} & 8 (8) \cellcolor{red!50} & 7 (7) \cellcolor{red!35} & 4 (6) \cellcolor{green!7}\\\hline
WB  & 5 (2) \cellcolor{red!7} & 3 (4) \cellcolor{green!21} & 7 (1) \cellcolor{red!35} & 2 (3) \cellcolor{green!35} & 1 (5) \cellcolor{green!50} & 8 (6) \cellcolor{red!50} & 6 (8) \cellcolor{red!21} & 4 (7) \cellcolor{green!7}\\\hline
\hline
Average  & 4.4 (4.3) \cellcolor{green!1} & 2.6 (3.7) \cellcolor{green!26} & 6.0 (\underline{2.0}) \cellcolor{red!21} & 2.4 (3.3) \cellcolor{green!30} & \textbf{1.0} (5.4) \cellcolor{green!50} & 8.0 (7.4) \cellcolor{red!50} & 5.8 (3.7) \cellcolor{red!18} & 5.8 (6.3) \cellcolor{red!18}\\\hline

        \end{tabular}%
        }
        
  \end{center}
  \label{tab:mraeranks}
\end{table}
Something that jumps to the eye when observing these tables is that
our experiments lead to conclusions that are \emph{dramatically
different} from those drawn by [GS2016]. First, SLD now unquestionably
emerges as the best performer, while it was often ranked among the
worst performers in [GS2016]. Conversely, PCC was the winner on most
combinations (dataset, measure) in [GS2016], while our experiments
have shown it to be a bad performer. Other methods too see their
merits disconfirmed by our experiments; in particular, ACC and PACC
have climbed up the ranked list, while all other methods (especially
SVM(KLD)) have lost ground.

The reason for the different conclusions that these two batches of
experiments allow drawing is, in all evidence, the amounts of
distribution shift which the methods have had to confront in the two
scenarios. In the experiments of [GS2016] this shift was \blue{very}
moderate, since the only test sample used (which coincided with the
entire test set) usually displayed class prevalence values not too
different from the class prevalence values in the training set. This
is shown in the last column of
Table~\ref{tab:datasetscharacteristics}, where the shift between
training set and test set (expressed in terms of absolute error) is
reported for each dataset; shift values range between 0.0020 and
0.1055, with an average value across all datasets of 0.0301, which is
a very low value. In our experiments, instead, the quantification
methods need to confront class prevalence values that are sometimes
\emph{very} different from the ones in the training set; shift values
range between 0.0000 and 0.6666, with an average value across all
samples of 0.2350. This means that the quantification methods that
have emerged in our experiments are the ones that are robust to
possibly radical changes in these class prevalence values, while the
ones that had fared well in the experiments of [GS2016] are the
methods that tend to perform well merely in scenarios where these
changes are bland.

\begin{figure}[t]
  \includegraphics[width=\textwidth]{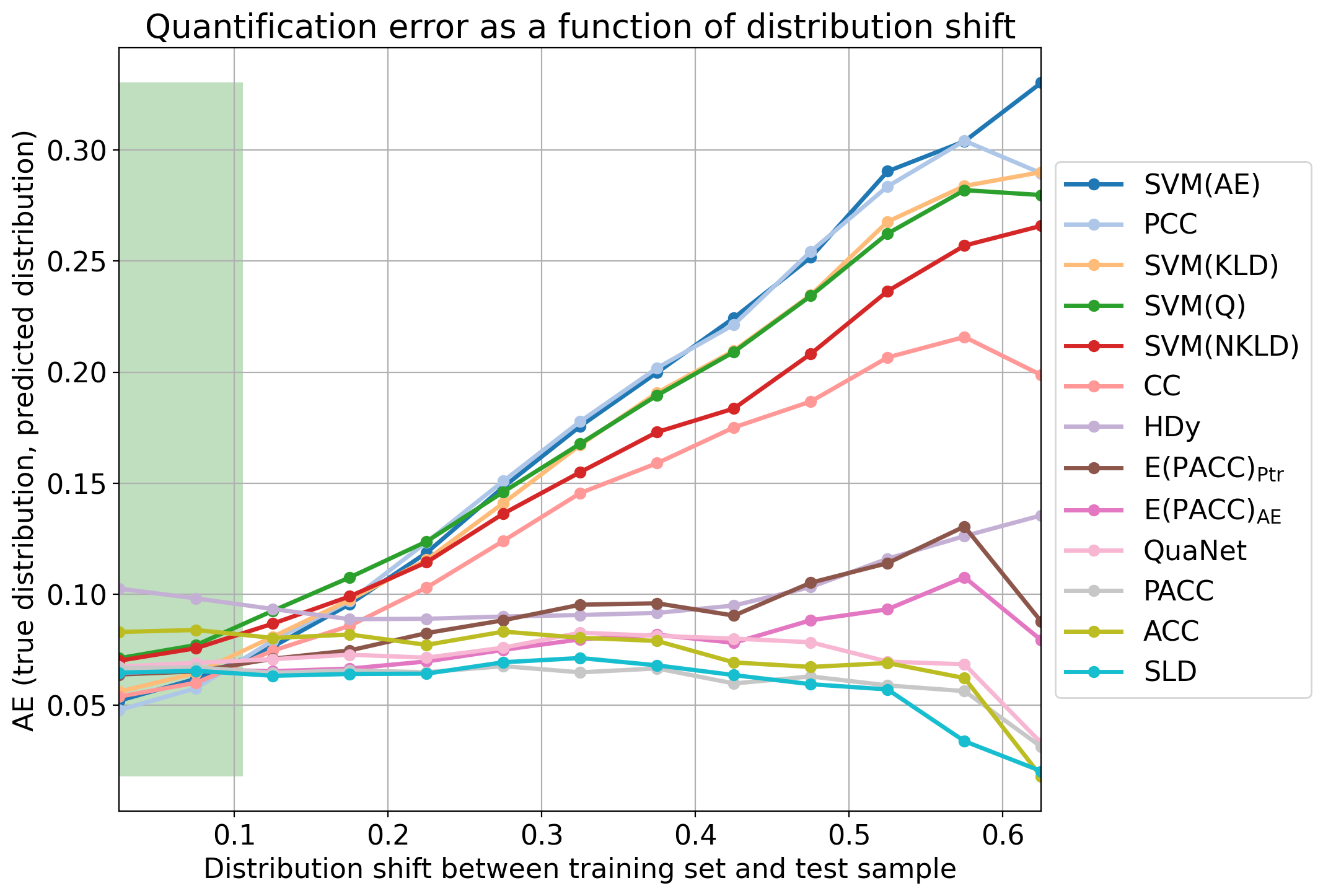}
  \caption{\label{fig:plotsae}Performance of the various
  quantification methods, \blue{represented by the coloured lines and}
  measured in terms of AE (lower is better), as a function of the
  distribution shift between training set and test sample; the results
  are averages across all samples \blue{in the same bin, i.e.,
  characterised by approximately the same amount of shift,
  independently of the dataset they were sampled from}. The \blue{two
  vertical dotted lines} indicate the range of distribution shift
  values exhibited by the experiments of [GS2016] \blue{(i.e., in
  those experiments, the AE values of distribution shift range between
  0.020 and 0.1055). The green histogram in the background shows
  instead how the samples we have tested upon are distributed across
  the different bins.}}

\end{figure}

\begin{figure}[t]
 \includegraphics[width=\textwidth]{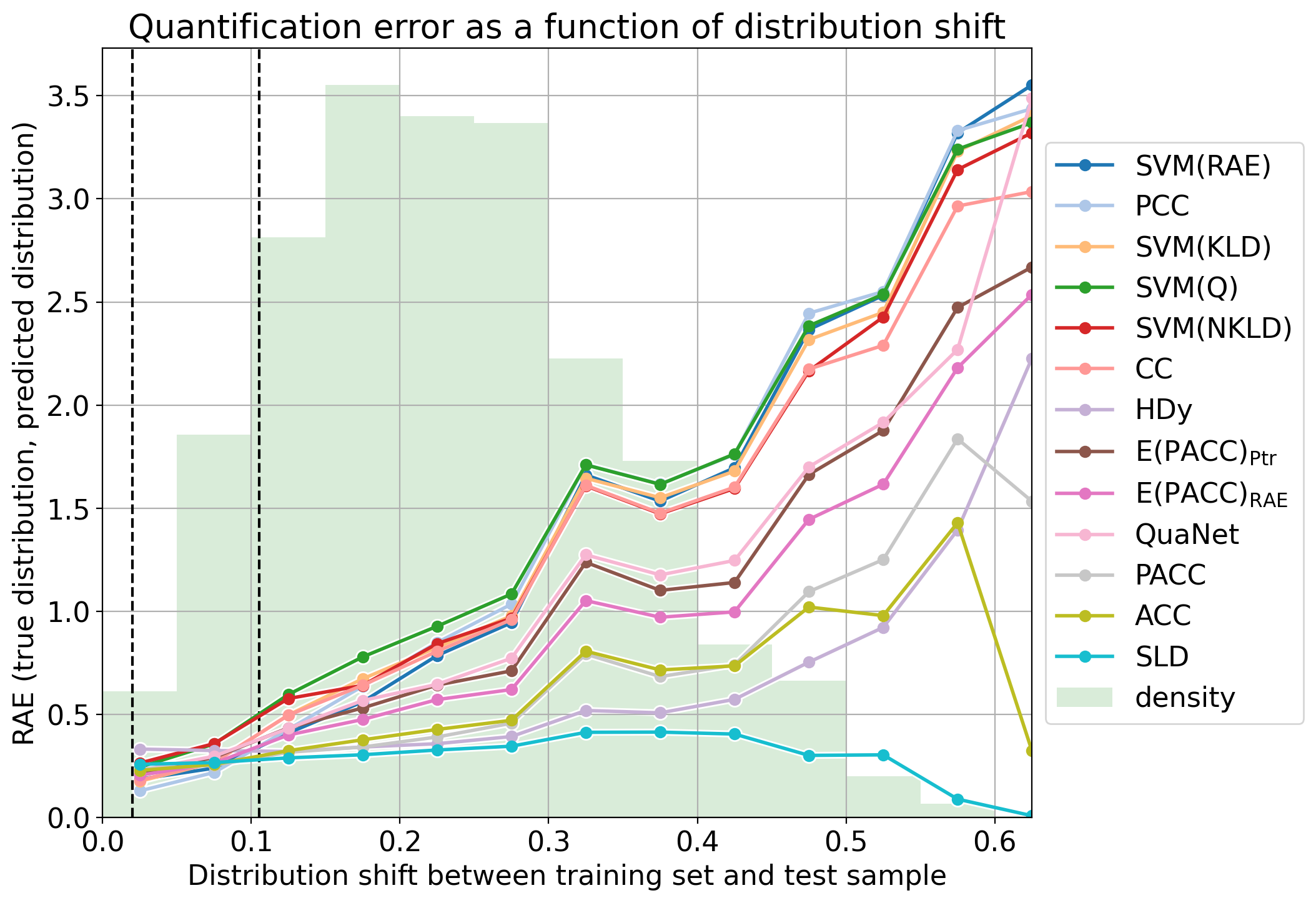}
  \caption{\label{fig:plotsrae}\blue{Performance of the various
  quantification methods, represented by the coloured lines and
  measured in terms of RAE (lower is better), as a function of the
  distribution shift between training set and test sample; the results
  are averages across all samples in the same bin, i.e., characterised
  by approximately the same amount of shift, independently of the
  dataset they were sampled from}. Unlike in Figure~\ref{fig:plotsae},
  for better clarity these results are actually displayed on a
  logarithmic scale.  \blue{The two vertical dotted lines indicate the
  range of distribution shift values exhibited by the experiments of
  [GS2016] (i.e., in those experiments, the AE values of distribution
  shift range between 0.020 and 0.1055). The green histogram in the
  background shows instead how the samples we have tested upon are
  distributed across the different bins.}}
\end{figure}

This situation is \blue{well} depicted in the plots of
Figures~\ref{fig:plotsae} and~\ref{fig:plotsrae}.
For generating these plots we have computed, for each of the
11$\times$\blue{5,775}=\blue{63,525} test samples, the distribution
shift between the training set and the test sample, and we have binned
these \blue{63,525} samples into bins characterised by approximately
the same amount of distribution shift (we compute distribution shift
as the absolute error between the training distribution and the
distribution of the test sample, using bins of width equal to 0.05
(i.e., [0.00,0.05], (0.05,0.10], etc.).
The plots show, for a given quantification method and for a given bin,
the quantification error of the method, measured (\blue{by means of}
AE in the top figure and \blue{by means of} RAE in the bottom figure)
as the average error across all samples in the same bin. \blue{The
green histogram in the background shows instead the distribution of
the samples across the bins. (See more on this at the end of this
section.)}

The plots clearly show that, for CC, PCC, SVM(KLD), 
SVM(NKLD), SVM(Q), as well as for the newly added SVM(AE) and
SVM(RAE), this error increases, in a very substantial manner as
distribution shift increases. A common characteristic of this group of
methods, that we will dub the ``unadjusted'' methods, is that none of
them attempts to correct, or adjust the counts
resulting from the classification of data items, thus resulting in
quantification systems that behave reasonably well for test set class
prevalence values close to the \blue{ones of the training set} (i.e.,
for low values of distribution shift), but that tend to generate large
errors for higher values of shift.
The obvious conclusion is that
failing to adjust makes the method not robust to high amounts of
distribution shift, and that the reason why some of
the unadjusted methods were successful in the evaluation of [GS2016]
is that this latter confronted the methods with very low amounts of
distribution shift. In fact, it is immediate to note from
Figures~\ref{fig:plotsae} and~\ref{fig:plotsrae} that, when
distribution shift is \blue{between 0.020 and 0.1055} (\blue{the
values} of distribution shift that the experiments of [GS2016] tackled
-- the region of Figures~\ref{fig:plotsae} and~\ref{fig:plotsrae}
\blue{between the two vertical dotted lines} encloses values of shift
up to that level), \blue{the difference in performance between
different quantification methods is small.}

In our plots, by contrast, methods ACC, PACC, SLD, along with the
newly added HDy, QuaNet, \EPACCAE, and \EPACCPtr, form a second group
of methods, that we will dub the ``adjusted'' methods, since they all
implement, in one way or \blue{another}, different strategies for
post-processing the class prevalence estimations \blue{returned} by
base classifiers.
The quantification error displayed by the ``adjusted'' methods remains
fairly stable across the entire range of distribution shift values,
which is clearly the reason of their success in the APP-based
evaluation we have presented
here.

\begin{figure}[t]
  \centering
  \includegraphics[width=0.9\textwidth]{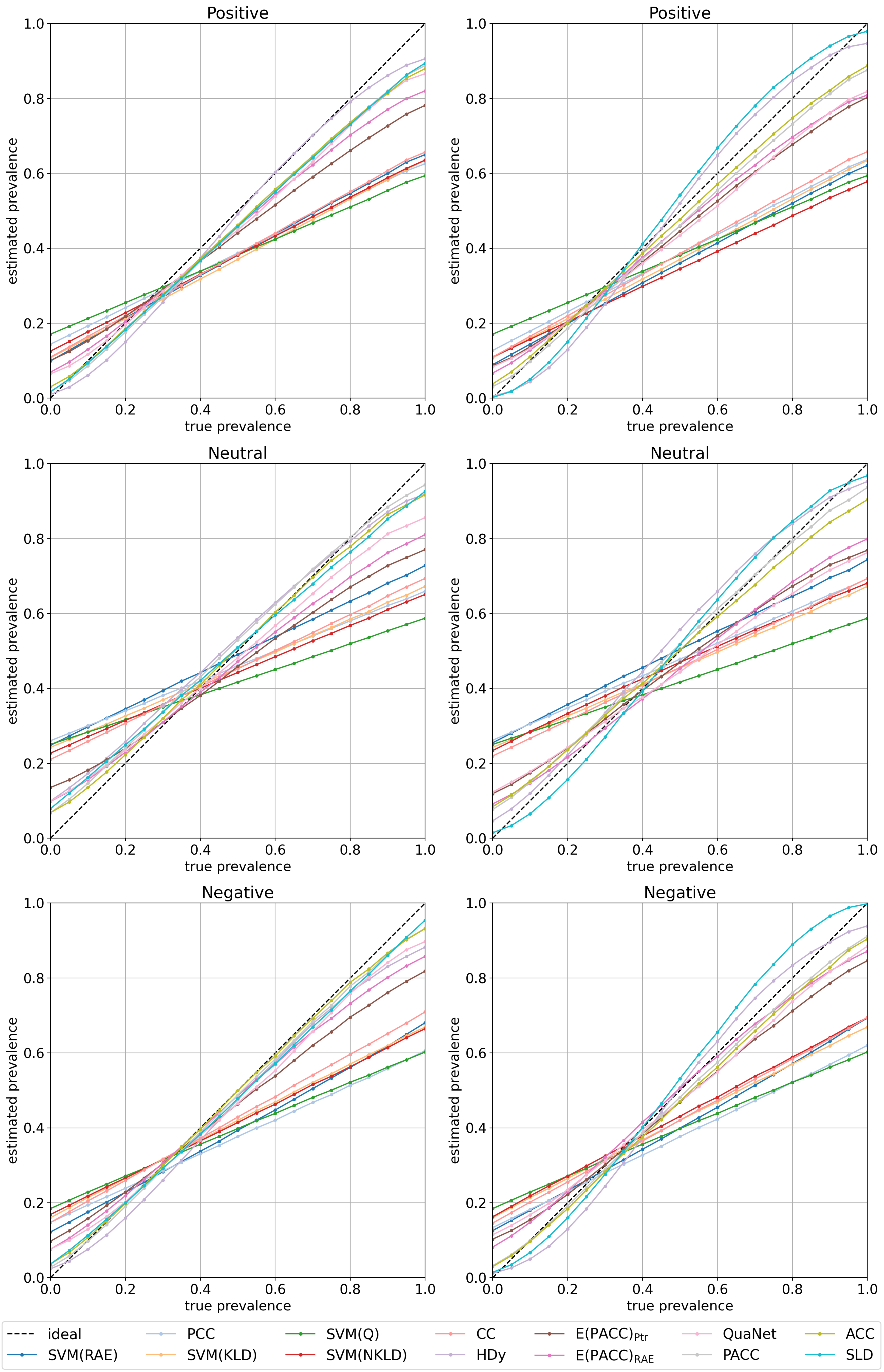}  
  \caption{Estimated prevalence as a function of true prevalence 
  according to various quantification methods. Results are displayed 
  separately for classes $\oplus$ (top), $\odot$ (middle), and 
  $\ominus$ (bottom), with methods optimized for according to AE (left) and RAE
  (right). 
  \label{fig:diag}}
\end{figure}

Figure~\ref{fig:diag} shows the estimated class prevalence value ($y$
axis) that each method delivers, on average across all test samples
and all datasets, for each true prevalence ($x$ axis); results are
displayed separately for each of the three target classes and for
methods optimized according to either AE or RAE. Note that the ideal
quantifier (i.e., one that makes zero-error predictions) would be
represented by the diagonal (0,0)-(1,1), here displayed as a dotted
line. These plots
support our observation that two groups of methods, the ``adjusted''
vs.\ the ``unadjusted'', exist (this is especially evident for the
$\oplus$ and the $\ominus$ classes, where they originate two quite
distinct bundles of curves), and show how the unadjusted methods fail
to produce good estimates for the entire range of prevalence
values. As could be expected, all methods intersect approximately in
the same point, which corresponds to the average training prevalence
of the class across all datasets
($p_L({\oplus})=0.278,p_L({\odot})=0.426,p_L({\ominus})=0.296$), given
that all methods tend to produce low error (hence similar values) for
test class prevalence values close to the training ones.

\begin{figure}[t]
  \centering
  \includegraphics[width=0.96\textwidth,keepaspectratio]{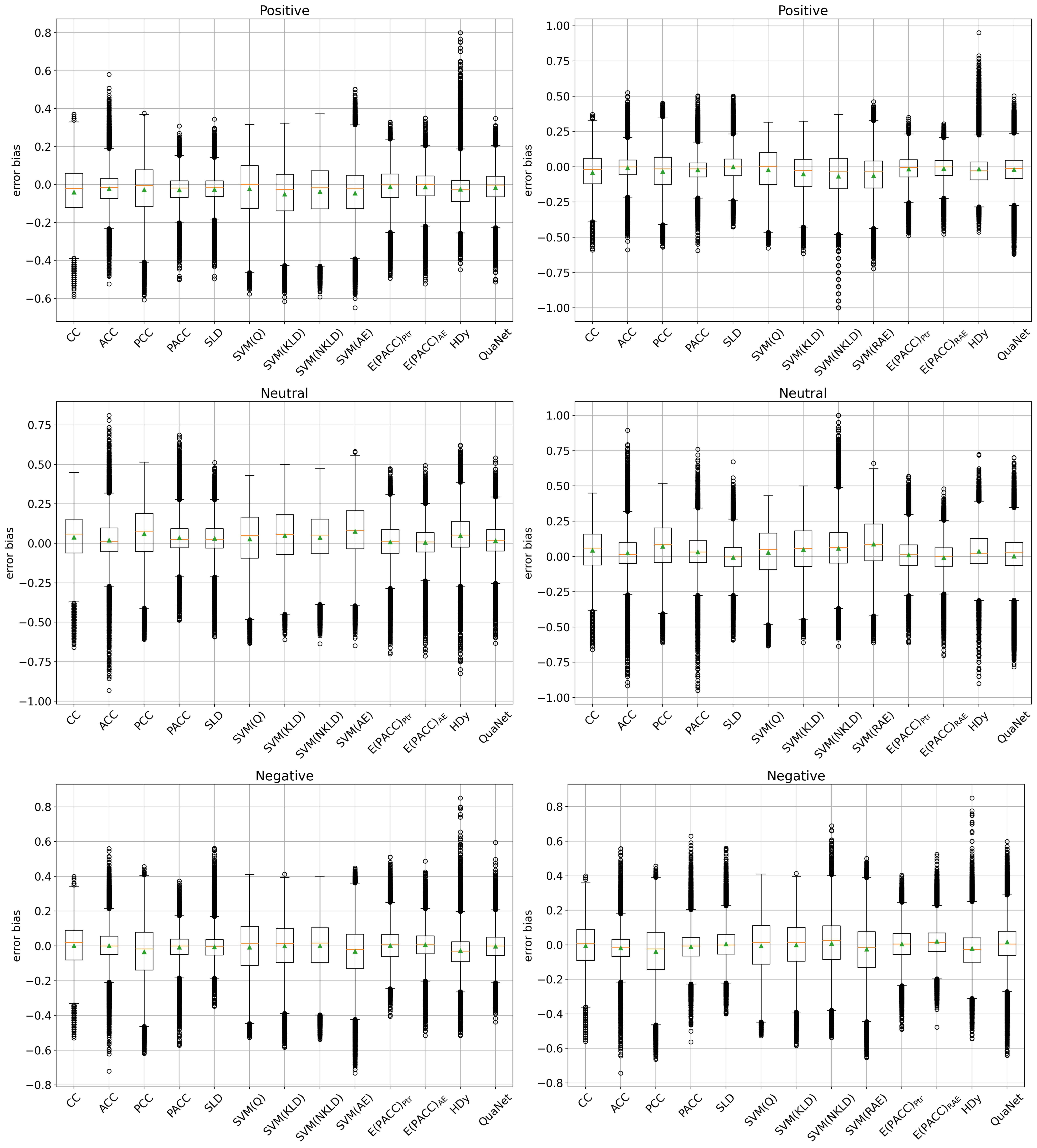}
  \caption{Box-plots of the error bias (signed error). Results are
  displayed separately for classes $\oplus$ (top), $\odot$ (middle),
  and $\ominus$ (bottom), with methods optimized for according to AE
  (left) and RAE (right).}
  \label{fig:boxplots}
\end{figure}

Figure~\ref{fig:boxplots} displays box-plot diagrams for the error
bias (i.e., for the signed error between the estimated prevalence
value and the true prevalence value) for all methods and independently
for each class, as averaged across all datasets and test samples. The
``adjusted'' methods show lower error variance, as witnessed by the
fact that their box-plots (indicating the first and third quartiles of
the distribution) tend to be squashed and their whiskers (indicating
the maximum and minimum, disregarding outliers) tend to be
shorter. Some methods tend to produce many outliers (see, e.g., ACC
and PACC in the $\odot$ class), which might be due to the fact that
the adjustments that those methods perform may become unstable in some
cases.\footnote{This instability is well known in the literature, and
has indeed motivated the appearance of dedicated methods that counter
the numerical instability that some adjustments may produce in the
binary case (see, e.g., \cite{Forman:2006uf,Forman:2008kx}).}
Overall, PACC and SLD, the two strongest methods among the
quantification systems we have tested, seem to be also the methods
displaying the smallest bias across the three classes.

As a final note, the reader might wonder why, for certain
well-performing methods, quantification error even seems to
\emph{decrease} for particularly high values of distribution shift
(see e.g., ACC, PACC, SLD in Figure~\ref{fig:plotsae} or SLD and ACC
in Figure~\ref{fig:plotsrae}). The answer is that quantification error
values for very high levels of shift are, in our experiments, not
terribly reliable, because \blue{(as clearly shown by the green
histograms in Figures~\ref{fig:plotsae} and~\ref{fig:plotsrae})} they
are averages across \emph{very few} data points. To see this, note
that the values of AE range (see \cite{Sebastiani:2020qf}) between 0
(best) and
\begin{equation}\label{eq:min}
  \begin{aligned}
    & \frac{2(1-\min_{y\in\mathcal{Y}}p(y))}{|\mathcal{Y}|}
  \end{aligned}
\end{equation}
\noindent (worst), which in our ternary case means
$\frac{2}{3}(1-0)=0.\overline{6}$ (because we indeed have test samples
in which the prevalence of at least one class is 0). However, there
are many more samples with extremely low AE values than samples with
extremely high AE values; for instance, out of the
$11\times$\blue{5,775}=\blue{63,525} samples that we have generated in
our experiments (see Section~\ref{sec:appnpp}), there are only 25
whose value of distribution shift is comprised in the interval
$[0.60, 0.6\overline{6}]$, while there are no fewer than \blue{3,300}
whose value is comprised in the interval $[0.00, 0.0\overline{6}]$,
even if the two intervals have the same width.
To see why, note for instance that we can reach an AE value of
$0.\overline{6}$ only when one of the classes in the training set has
a prevalence value of 0 (see Equation~\ref{eq:min}), while an AE value
of 0 can be reached for all training sets. As a result, the average AE
values at the extreme right of the plots in Figures~\ref{fig:plotsae}
and~\ref{fig:plotsrae} (say, those beyond $x=0.55$) are averages
across very few data points, and are thus unstable and
unreliable. This does not invalidate our general observations, though,
since each quantification method we test displays, on the [0.00,0.55]
interval, a very clear, unmistakable behaviour.


\section{Conclusions}
\label{sec:conclusions}

\noindent The results of our experiments show that a re-evaluation of
the relative merits of different quantification methods on the tweet
sentiment quantification task was necessary. We have shown that the
\blue{experimentation} previously conducted in [GS2016] was
\blue{weak, since the experimental protocol that was followed} led the
authors of this study to conduct their evaluation on a radically
insufficient amount of test data points. We have then conducted a
re-evaluation of the same methods on the same datasets according to a
more robust, and now \blue{widely} accepted, experimental protocol,
which has lead to an experimentation \blue{on a number of datapoints
5,775} times larger than the one of [GS2016]. \blue{In addition to
these experiments}, we have also tested some \blue{further} methods,
some of which had appeared after [GS2016] was published.

This experimentation has proven necessary for at least two
reasons. The first reason is that some of the
evaluation functions (such as KLD and NKLD) that had been used in
[GS2016] are now known to be unsatisfactory, and their use should thus
be deprecated in favour of functions such as AE and RAE. The second
reason, and probably the most important one, is that the results of
our re-evaluation have radically disconfirmed the conclusions
originally drawn by the authors of [GS2016]\blue{, showing that the
methods (e.g., PCC) who had emerged as the best performers in [GS2016]
tend to behave well only in situations characterised by very low
distribution shift; on the contrary, when distribution shift
increases, other methods (such as SLD) are to be preferred.} In
particular, our experiments do justice to the SLD method, which had
obtained fairly bland results in the experiments of [GS2016], and
which now emerges as the true leader of the pack\blue{, thanks to
consistently good performance across the entire spectrum of
distribution shift values.}


\section*{Acknowledgments}
\label{sec:Acknowledgments}

\noindent 
The present work has been supported
by the \textsc{SoBigData++} project, funded by the European Commission
(Grant 871042) under the H2020 Programme INFRAIA-2019-1, and by the
\textsc{AI4Media} project, funded by the European Commission (Grant
951911) under the H2020 Programme ICT-48-2020. The authors' opinions
do not necessarily reflect those of the European Commission.



\end{document}